\documentclass[sigconf]{acmart}
\renewcommand\footnotetextcopyrightpermission[1]{} % Removes footnote with conference info
\AtBeginDocument{%
  }

\usepackage{amsmath,amsfonts}
\usepackage{array}
\usepackage{textcomp}
\usepackage{stfloats}
\usepackage{url}
\usepackage{verbatim}
\usepackage{graphicx}

\usepackage{algorithm}
\usepackage{algpseudocode}

\usepackage{multirow}
\usepackage{color}
\usepackage{caption}
\usepackage{url}
\usepackage{booktabs}
\usepackage{diagbox}
\usepackage{subcaption}
\usepackage{bbm}
\usepackage{dsfont}

\begin{document}

%%
%% The "title" command has an optional parameter,
%% allowing the author to define a "short title" to be used in page headers.
\title{TeSO: Representing and Compressing 3D Point Cloud Scenes with Textured Surfel Octree}

%%
%% The "author" command and its associated commands are used to define
%% the authors and their affiliations.
%% Of note is the shared affiliation of the first two authors, and the
%% "authornote" and "authornotemark" commands
%% used to denote shared contribution to the research.

% \author{Yueyu Hu, Ran Gong, Tingyu Fan, and Yao Wang}
% \affiliation{%
%   \institution{Dept. Electrical and Computer Engineering, Tandon School of Engineering, New York University}
%   \city{Brooklyn}
%   \country{USA}}
% \email{{yyhu,rg4827,tingyufan,yaowang}@nyu.edu}

\author{Yueyu Hu}
\affiliation{%
  \institution{Dept. Electrical and Computer Engineering, Tandon School of Engineering, New York University}
  \city{Brooklyn}
  \country{USA}}
\email{yyhu@nyu.edu}

\author{Ran Gong}
\affiliation{%
  \institution{Dept. Electrical and Computer Engineering, Tandon School of Engineering, New York University}
  \city{Brooklyn}
  \country{USA}}
\email{rg4827@nyu.edu}

\author{Tingyu Fan}
\affiliation{%
  \institution{Dept. Electrical and Computer Engineering, Tandon School of Engineering, New York University}
  \city{Brooklyn}
  \country{USA}}
\email{tingyufan@nyu.edu}

\author{Yao Wang}
\affiliation{%
  \institution{Dept. Electrical and Computer Engineering, Tandon School of Engineering, New York University}
  \city{Brooklyn}
  \country{USA}}
\email{yaowang@nyu.edu}

%%
%% By default, the full list of authors will be used in the page
%% headers. Often, this list is too long, and will overlap
%% other information printed in the page headers. This command allows
%% the author to define a more concise list
%% of authors' names for this purpose.
\renewcommand{\shortauthors}{Hu et al.}

%%
%% The abstract is a short summary of the work to be presented in the
%% article.
\begin{abstract}
  
3D visual content streaming is a key technology for emerging 3D telepresence and AR/VR applications. One fundamental element underlying the technology is a versatile 3D representation that is capable of producing high-quality renders and can be efficiently compressed at the same time. Existing 3D representations like point clouds, meshes and 3D Gaussians each have limitations in terms of rendering quality, surface definition, and compressibility. In this paper, we present the Textured Surfel Octree (TeSO), a novel 3D representation that is built from point clouds but addresses the aforementioned limitations. It represents a 3D scene as cube-bounded surfels organized on an octree, where each surfel is further associated with a texture patch. By approximating a smooth surface with a large surfel at a coarser level of the octree, it reduces the number of primitives required to represent the 3D scene, and yet retains the high-frequency texture details through the texture map attached to each surfel. We further propose a compression scheme to encode the geometry and texture efficiently, leveraging the octree structure. The proposed textured surfel octree combined with the compression scheme achieves higher rendering quality at lower bit-rates compared to multiple point cloud and 3D Gaussian-based baselines.

\end{abstract}

%%
%% The code below is generated by the tool at http://dl.acm.org/ccs.cfm.
%% Please copy and paste the code instead of the example below.
%%
% \begin{CCSXML}
% <ccs2012>
%    <concept>
%        <concept_id>10002951.10003227.10003251.10003255</concept_id>
%        <concept_desc>Information systems~Multimedia streaming</concept_desc>
%        <concept_significance>300</concept_significance>
%        </concept>
%    <concept>
%        <concept_id>10010147.10010371.10010396.10010399</concept_id>
%        <concept_desc>Computing methodologies~Parametric curve and surface models</concept_desc>
%        <concept_significance>300</concept_significance>
%        </concept>
%    <concept>
%        <concept_id>10002951.10002952.10002971.10003451.10002975</concept_id>
%        <concept_desc>Information systems~Data compression</concept_desc>
%        <concept_significance>300</concept_significance>
%        </concept>
%    <concept>
%        <concept_id>10010147.10010371.10010396.10010400</concept_id>
%        <concept_desc>Computing methodologies~Point-based models</concept_desc>
%        <concept_significance>300</concept_significance>
%        </concept>
%  </ccs2012>
% \end{CCSXML}

% \ccsdesc[300]{Information systems~Multimedia streaming}
% \ccsdesc[300]{Computing methodologies~Parametric curve and surface models}
% \ccsdesc[300]{Information systems~Data compression}
% \ccsdesc[300]{Computing methodologies~Point-based models}

%%
%% Keywords. The author(s) should pick words that accurately describe
%% the work being presented. Separate the keywords with commas.
\keywords{3D Representation, Textured Surfel Octree, 3D Gaussians, Point Cloud Compression}

% \received{20 February 2007}
% \received[revised]{12 March 2009}
% \received[accepted]{5 June 2009}

%%
%% This command processes the author and affiliation and title
%% information and builds the first part of the formatted document.
\maketitle

\section{Introduction}\label{sec:intro}

3D telepresence is a promising technology that can revolutionize remote communication, with emerging applications such as remote AR education and collaboration~\cite{funk2017holocollab,kaminska2023augmented,reipschlager2019designar}, 3D video conferencing~\cite{lawrence2024project,stengel2023ai,jin20243d}) and immersive gaming experiences. A key challenge in 3D telepresence is the high data volume of 3D visual contents, especially for applications that require real-time high-quality streaming and interaction via the Internet.  An efficient 3D representation and compression techniques for the representation are essential to facilitate 3D telepresence~\cite{guan2023metastream,li2024spatial}.

\begin{figure}
    \centering
    \includegraphics[width=1.\linewidth]{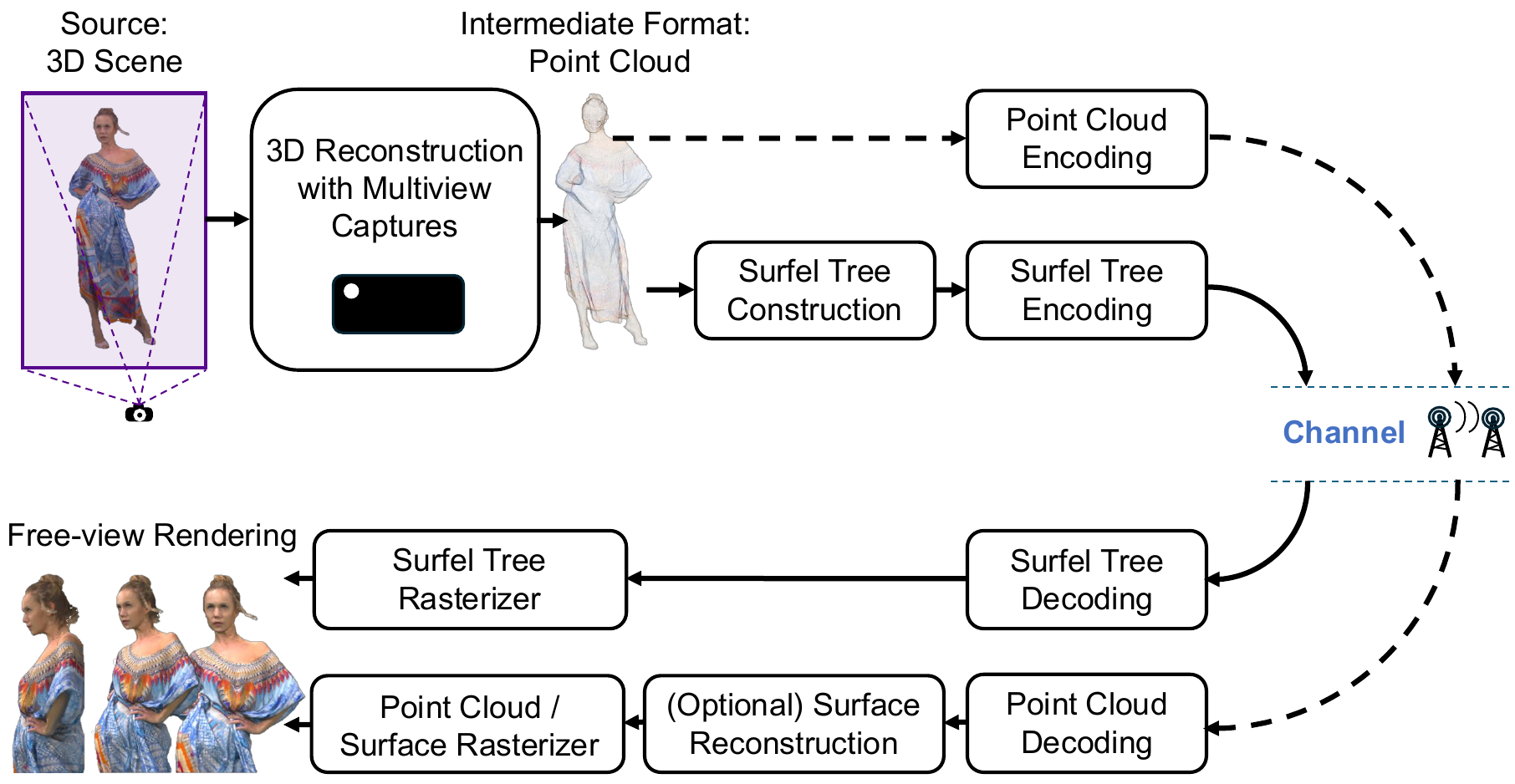}
    \caption{Comparison of the proposed TeSO in a 3D streaming system with a standard point cloud-based pipeline.}
    \label{fig:system}
\end{figure}

To this end, point clouds have been widely adopted as a promising 3D representation~\cite{liang2024fumos,lee2020groot,liu2023cav3,zhang2022yuzu}, because they are flexible for various 3D scenes and are amenable for compression and processing. The importance is also evidenced by the MPEG point cloud compression standardization activities since 2014~\cite{chen2023introduction,gpccwhitepaper}. Point clouds have advantages over conventional meshes in compactness as we do not need to describe the vertex connectivity information~\cite{zerman2020textured}. It also has high flexibility, \textit{e.g.} it can be efficiently downsampled for coding at a lower bit-rate. However, due to the loss of connectivity and the lack of explicit surface representation, to render point clouds without holes from arbitrary views, we need to either conduct surface reconstruction and convert point clouds into meshes~\cite{zerman2020textured}, or render with more advanced machine learning-based methods~\cite{hasegawa2024blendpcr,hu2024low,chang2023pointersect}. These additional processing steps can be time- and energy-consuming, making them not suitable for real-time applications on head-mounted devices. Besides, since point clouds restrict the color specification to a single color for each point, when the number of points is reduced, the texture rendering quality is inevitably degraded. This is especially problematic for areas that require high-frequency texture details, such as human faces and clothing. 

To address these limitations of point cloud as the 3D representation in the streaming scenarios, in this paper, we present the Textured Surfel Octree (TeSO), a novel 3D representation that combines efficiency of point clouds and explicit surface definition of meshes. The key idea is to represent the surface of a 3D scene as \textbf{octree cubes-bounded textured surfels} (\textit{i.e.}, \textbf{surf}ace \textbf{el}ements). The surfels explicitly describe the surface as in meshes, but without the burden of specifying vertex connectivity information. Hence, it combines the advantages of mesh and point-based rendering primitives like 3D Gaussians, offering surface representation and flexibility. With the proposed CUDA-based rasterizer, it supports real-time rendering with six degrees of freedom (6 DoF). Compared to 3D Gaussians, it can reduce the number of primitives in smooth areas, to trade off rendering quality for reduced bandwidth consumption and computational complexity, while maintaining the texture details. This allows TeSO to deliver uncompromising rendering quality even when the geometry is simplified. 

We further propose a coding scheme to directly compress TeSO. As shown in Fig.~\ref{fig:system} where we compare the proposed pipeline with a standard point-cloud-based pipeline (in dashed lines), the standard system needs a surface reconstruction or a sophisticated renderer to achieve high-quality and hole-free rendering, whereas the proposed scheme provides high rendering quality without the need for decoder-side surface reconstruction. With this scheme, we can achieve efficient decoding and rendering of the 3D scene stream, facilitating the broad adoption of 3D streaming applications. We will open-source the TeSO construction, compression and rendering implementation. The main contributions of this work are summarized as follows:
\begin{itemize}
    \item We propose the textured surfel octree, a novel 3D representation that combines the advantages of point clouds and meshes. TeSO can represent explicit surface without specifying additional connectivity information, making it capable of achieving gap-free high-quality rendering while being efficient for compression. 
    \item We present a construction algorithm to build TeSO from point clouds. Since the algorithm is designed for and implemented with GPU acceleration, we achieve fast surfel octree building from point cloud in less than 0.5 seconds.
    \item We propose a compression scheme for TeSO based on a learned convolutional entropy model for the geometry and a standard codec for point clouds or images. With the proposed compression scheme, we show that TeSO can achieve superior rendering quality at similar bit-rates compared to existing approaches based on G-PCC \cite{gpccwhitepaper}, learned point cloud rendering~\cite{hu2024low} and the end-to-end point cloud compression-rendering codec~\cite{hu2024bits}.
\end{itemize}

The remainder of this paper is organized as follows. In Section~\ref{sec:prior}, we review related work in point cloud compression, rendering, and various 3D content representations. In Section~\ref{sec:surfel_tree}, we present the structure and construction of TeSO. In Section~\ref{sec:rasterization}, we describe the efficient rendering method. In Section~\ref{sec:compression}, we present the proposed compression scheme. Finally, we present experimental results comparing our approach with several baselines in terms of the rate-distortion performance and visual quality in Section~\ref{sec:result} and conclude the paper in Section~\ref{sec:conclusion}.

\section{Related Work}\label{sec:prior}

\subsection{Point Cloud Compression}\label{sec:pcc}

With the wide potential applications of point clouds for VR / AR, immersive streaming, and interactive 3D telepresence, which requires transmission and storage of point clouds, point cloud compression has been an active research area. The MPEG 3D Graphics and Haptics Coding group has standardized the Video-based Point Cloud Compression~(V-PCC)~\cite{tmc2,graziosi2020overview} and Geometry-based Point Cloud Compression~(G-PCC) standards~\cite{gpccwhitepaper,tmc13}. Beyond that, various learning-based point cloud codecs have been proposed~\cite{huang2020octsqueeze,que2021voxelcontext,fu2022octattention,cui2023octformer,mao2022learning,wang2021lossy,wang2022sparsepcgc,he2022density,zhang2023yoga,wang2024versatile,wang2024versatile2}, with some demonstrating better rate-distortion performances in terms of the geometry accuracy and color fidelity than G-PCC.

While most of the existing work aims to minimize point cloud geometry and color distortion, B2P~\cite{hu2024bits} proposes to directly optimize point cloud compression towards the trade-off between the rendering distortion versus bit-rate. To achieve this goal, B2P encodes a point cloud into a neural feature bit-stream, which is decoded directly to 3D Gaussians for differentiable rendering. Since the 3D Gaussians representation specifies only one color value\footnote{Although by incorporating Spherical Harmonics with 3D Gaussians, one Gaussian can represent different color from different viewing directions~\cite{kerbl20233d}, for a given camera view, one Gaussian can only show a single color. Therefore, Spherical Harmonics cannot be used to represent spatially varying texture on a single Gaussian.}  for each Gaussian, it is not possible to represent high-frequency texture details when the number of Gaussians is restricted due to bit-rate or complexity constraints. In this work, we propose to overcome this limitation by specifying a high-frequency texture on each of the primitives in TeSO, which allows us to represent high-frequency texture details even when the geometry is simplified. We show that TeSO achieves better rendering quality than 3D Gaussians generated and compressed through B2P at the same bit-rate budget.

\subsection{Point Cloud Rendering}\label{sec:pcr}
Rendering point clouds into high-quality 2D images for visual display has been a long-standing challenge, because point clouds do not have explicitly defined surface. To address this issue, existing point-cloud rendering methods usually need to assign a splatting area to each point, and then render the point cloud by rasterizing the splats on the image plane, effectively converting points into a different graphical primitive. Typical splats include solid camera plane aligned squares used in Open3D / OpenGL~\cite{zhou2018open3d}, Gaussians with multivariate covariance matrices~\cite{zwicker2004perspective,zwicker2001ewa,zwicker2001surface}, and surfels~\cite{pfister2000surfels}. The splatting-based rendering methods do not require a complex surface reconstruction process to build the explicit connectivity between points, and therefore are generally efficient.

However, these point primitive-based approaches handle each point individually, without forming a surface. This leads to visibility issues (\textit{i.e.} non-visible primitives are not completely occluded by closer ones). To address this visibility issue, splatting-based rendering methods either conduct a visibility check for each pixel~\cite{kobbelt2004survey}, or make the splats heavily overlap with each other so the gaps are not visually noticeable, which leads to blurry renders and inaccurate boundaries. Although per-scene optimization over the multivariate covariance matrices (as done in 3D Gaussian splatting~\cite{kerbl20233d}) can alleviate the blurry problem by fitting particular rendered views, because no explicit surface is formed, the optimization results do not generalize to various camera settings (\textit{e.g.} focal length, resolutions, and camera-object distance).

With the advancement in deep learning, there have been works employing neural networks to improve point cloud rendering. In \cite{hu2024low}, a sparse convolution-based neural network is proposed to convert each point in the point cloud into a 3D Gaussian, by estimating additional parameters such as the covariance matrix and center coordinate offset. Other primitives like neural points (points with attached neural features) can also be used~\cite{aliev2020neural}, where a 2D convolutional network is employed for converting feature map splats into rendered images. Instead of converting points to a different primitive, \cite{chang2023pointersect} proposes  a transformer based neural network  to directly calculate the intersection between a ray and the underlying implicit surface represented by the point cloud. However, it requires running a heavy transformer for each pixel with multiple points within the cylinder of a pixel ray, and incurs high computational cost and latency, making them unsuitable for real-time applications.

In this work, we propose TeSO as a new 3D representation,  which can serve as a method for point cloud rendering. In addition, we design this representation with compression compactness in mind, and show the strong capability of the representation to render high-quality images under bit-rate constraints.

\subsection{Other 3D Visual Content Representations and their Compression}\label{sec:3dvc}

Recent years have witnessed the proliferation of various 3D representations for modeling 3D visual contents, including Signed / Unsigned Distance Field (SDF / UDF)~\cite{lindell2022bacon,chibane2020neural,zhou2024cap,li2024learning}, Neural Radiance Field (NeRF)~\cite{mildenhall2021nerf}, 3D Gaussians~\cite{kerbl20233d}, and their variations~\cite{tewari2022advances}, along with efforts in developing compression methods for these representations~\cite{bird20213d,niedermayr2024compressed,takikawa2022variable,lionar2023nu}. However, these 3D representations are usually not designed in the first place for the purpose of streaming and with compression compactness in mind, and therefore are limited by nature in terms of regularity in structure (\textit{i.e.} primitives are not organized on a grid or other regular data structure) and in rendering efficiency (\textit{e.g.} NeRF and SDF cannot be rasterized directly). Our proposed TeSO is designed to be efficient for compression, and to afford direct rendering  without further post-processing. Therefore, it can serve as a good candidate representation for 3D visual content streaming.

\section{Structure and Construction of TeSO}\label{sec:surfel_tree}

\subsection{Motivation}

\begin{figure}[t]
    \centering
    \begin{subfigure}{0.3\linewidth}
        \centering
        \includegraphics[width=0.99\linewidth]{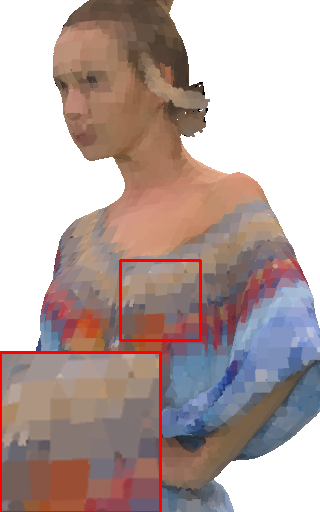}
        \caption{Average Color}
    \end{subfigure}
    \begin{subfigure}{0.3\linewidth}
        \centering
        \includegraphics[width=0.99\linewidth]{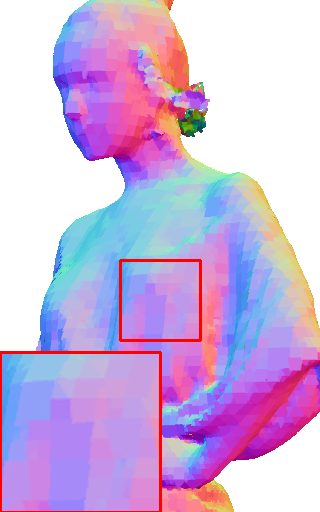}
        \caption{Surface Normal}
    \end{subfigure}
    \begin{subfigure}{0.3\linewidth}
        \centering
        \includegraphics[width=0.99\linewidth]{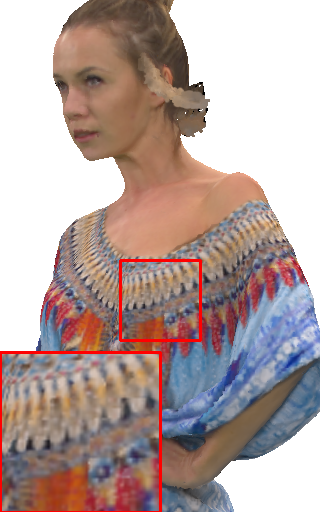}
        \caption{Texture}
    \end{subfigure}

    \caption{Illustration of geometry and texture complexity. (a) Each surfel shows the average color. (b) Visualization of surface normals on the surfels. (c) Rendering results with texture. As shown in the blow-up area, while the geometry is smooth and can be represented by a few surfel primitives, the texture is complex and requires high resolution to represent.}
    \label{fig:complexity}
\end{figure}

Unlike 2D images and videos where pixels are captured and organized on a 2D regular grid, 3D visual contents are usually represented as two parts, \textit{i.e.} geometry and color, due to the nature of sparsity in the 3D space and the fact that multi-camera capturing setup and LiDAR systems only capture the surface of a 3D scene. The geometry and color may have different complexity in different regions of the scene. An example illustrating this observation is shown in Fig.~\ref{fig:complexity}. Intuitively, for areas with smooth or even flat surfaces, we can use fewer graphical primitives to represent the geometry, leading to higher representation and rendering efficiency. However, these areas may contain complex textures, which require higher resolution to represent the color. In the context of lossy compression, in circumstances where we have a tight bit-rate budget, we can opt for a lower resolution for the geometry, while this should not be at the cost of reduced texture resolution. Besides, even when we trade off the geometry accuracy to bit-rate saving, we still need to ensure that the geometry parameters still represent the surface and be gap-free, in order to avoid severe artifacts in the rendering. In other words, we prefer a coarser and smoother geometry to a wrong geometry.

\subsection{Surfel Octree Geometry}\label{sec:geometry}

\begin{figure*}[t]
    \centering
    \begin{subfigure}{0.23\linewidth}
        \centering
        \includegraphics[width=0.9\linewidth]{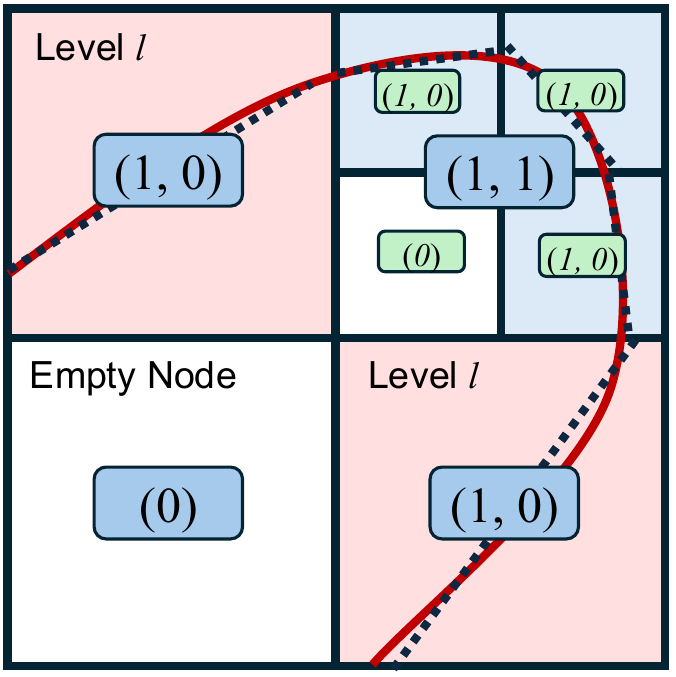}
        \caption{Cube bounded surfels.}\label{fig:cube}
    \end{subfigure}
    \begin{subfigure}{0.265\linewidth}
        \centering
        \includegraphics[width=0.9\linewidth]{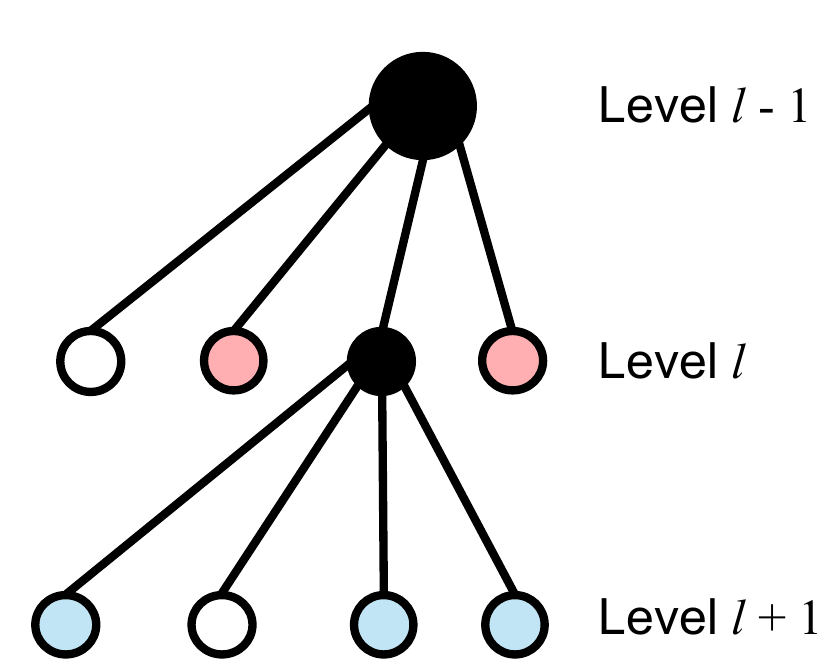}
        \vspace{1.6mm}
        \caption{Octree}\label{fig:octree}
    \end{subfigure}
    \begin{subfigure}{0.26\linewidth}
        \centering
        \includegraphics[width=0.85\linewidth]{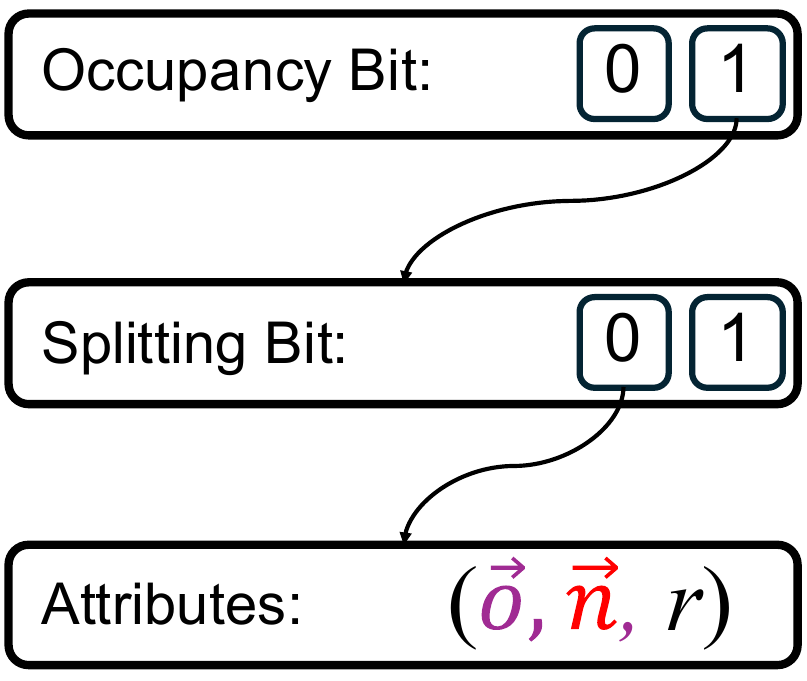}
        \vspace{2mm}
        \caption{Specified flags.}\label{fig:flags}
    \end{subfigure}
    \begin{subfigure}{0.22\linewidth}
        \centering
        \includegraphics[width=0.9\linewidth]{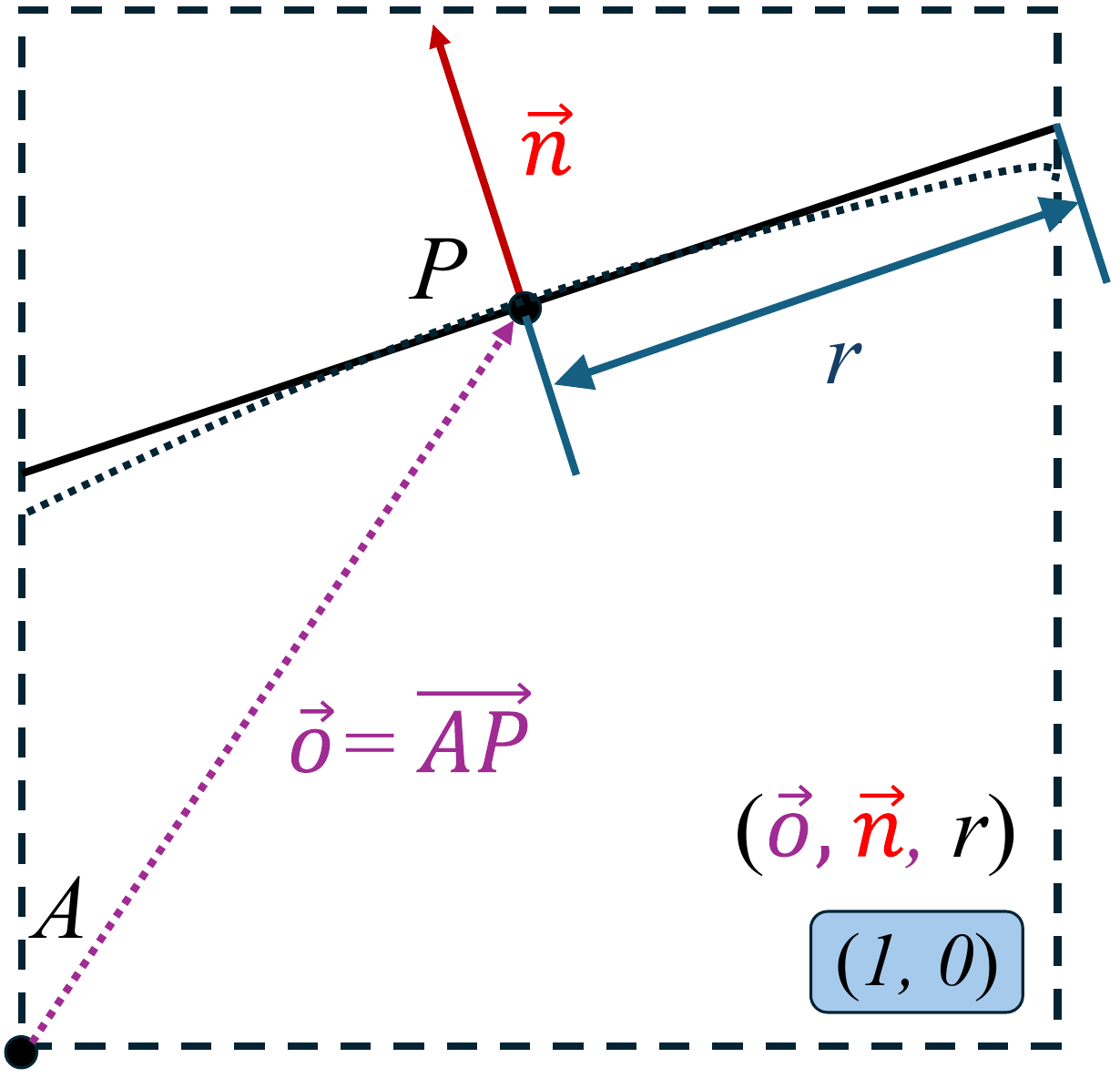}
        \vspace{1.5mm}
        \caption{Geometry attributes.}\label{fig:geometry_attributes}
    \end{subfigure}
    \caption{2D illustration of the structure of the surfel octree. When approximating an underlying surface (solid curve in red), areas that are mostly flat are represented using a coarse surfel at a lower octree level (pink nodes). If a coarse surfel cannot accurately represent the underlying surface, we split the cube into smaller cubes (blue nodes) and assign a surfel to each of them.}
    \label{fig:design}
\end{figure*}

The observation of geometry and texture complexity leads us to the design of the surfel octree representation. For the geometry in 3D scenes, we use a plane (\textit{i.e.} surfel) to approximate the geometry of an area of smooth surface, and additionally attach a patch of texture map to represent the color in this local area. As illustrated in Fig.~\ref{fig:cube}, we represent a smooth surface using a set of surfels. These surfels are organized on the leaf nodes of an octree, forming the \textbf{surfel octree} representation. We take the advantage of the hierarchical structure of an octree, which allows splitting a large cube into eight octants (sub-cubes) when necessary. For surface areas with nearly flat geometry, we can use a single surfel to represent the geometry (denoted level $l$ in Fig.~\ref{fig:cube} and Fig.~\ref{fig:octree}). For areas that have larger curvature or more complex texture, we further split the octree node into children nodes (denoted level $l+1$), and assign a surfel to each of the non-empty children nodes, if the surface in that node is flat. Otherwise, that node would be further split. As shown in Fig.~\ref{fig:flags}, for each node in level $l$ in the octree, we first use a 1-bit flag (\textit{i.e. occupancy bit}) to indicate its occupancy. If the node is occupied, we use another 1-bit  to indicate whether the node is a leaf node or not. If the node is a leaf node, we assign a surfel to it. The surfel is defined by geometry attributes $(\vec{o}, \vec{n}, r)$, denoting the offset vector from an anchor point of the cube, the normal vector, and the radius of the surfel, respectively, as shown in Fig.~\ref{fig:geometry_attributes}.

\begin{figure}
    \centering
    \includegraphics[width=0.95\linewidth]{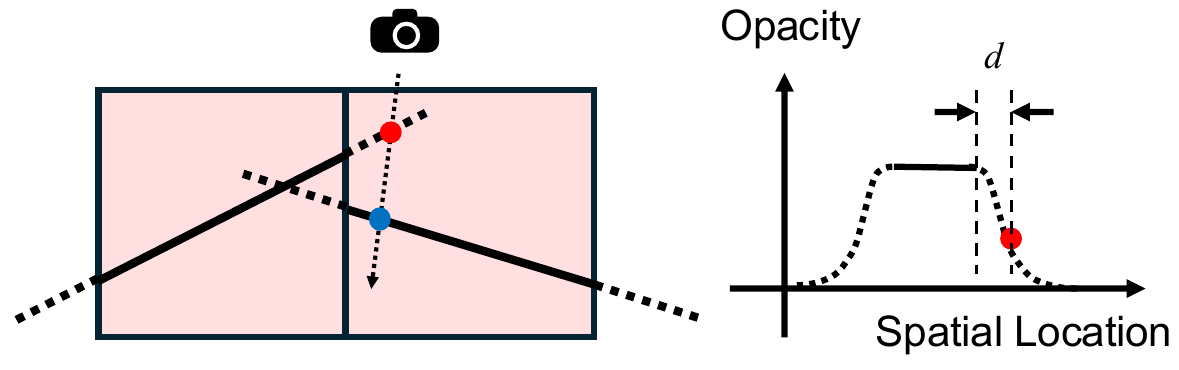}
    \caption{Illustration of the soft area extending outside the cube boundary for surfels.}
    \label{fig:soft}
\end{figure}

While the surfel octree representation reduces the number of primitives by simplifying the geometry, a problem arises as there is no regular 2D shape (\textit{e.g.} rectangle, ellipse) that can tightly represent a 3D surface without leaving gaps between the surfels or heavily overlapping with each other. To address this issue, we bound the surfel by the cube of the octree node, and use the cube boundary to define the splatting area of the surfel. This will produce polygonal cuts of the surfel inside the cube that closely connects to their neighbors, as we can observe in Fig.~\ref{fig:complexity}~(a) and (b). To further alleviate cracks in the rendering results due to misaligned boundaries in 3D, during rendering, we extend each surfel with a \textit{soft} area outside the bounding cube, as illustrated in Fig.~\ref{fig:soft}. The soft area on each surfel behaves like a translucent surface, which allows the surfel to blend with its neighbors. If the intersecting point of a ray on the surfel is within the soft area, the rendered color is obtained by blending with the neighboring surfels using $\alpha=e^{-\frac{d^2}{\sigma^2}}$, 
where $d$ is the shortest distance from the intersecting point to the bounding cube, and $\sigma$ is a hyperparameter that controls the blending range. We follow the same blending mechanism as used in 3D Gaussian Splatting~\cite{kerbl20233d}, except that virtual camera ray always stops at the first intersection with a solid surfel. We set $\sigma$ to be the width of the smallest bounding cube in the octree, which also serves as a texture filter when the granularity of the geometry reaches the finest level. This design ensures the smoothness of the rendering results and avoids gaps between surfels.

\subsection{Surfel Octree Construction from Point Cloud}

\begin{algorithm}
    \footnotesize
        \caption{Calculate Surfel Octree from Point Cloud}\label{alg:build}
        \begin{algorithmic}
        \Require Point Cloud $\{x_i\}$, Octree Levels $\{l_i\}$, Criteria $f()$
        \Ensure Surfel Octree Geometry $T = \{(\vec{o}_i, \vec{n}_i, r_i)\}$
        \State $UnvisitedPoints \gets \{x_i\}$
        \State $T \gets \emptyset$
        \State $\{l_i\} \gets \text{Sort}(\{l_i\})$ \Comment{Sort levels in ascending order}
        \For {$l$ in $\{l_i\}$}
            % \State $UnvisitedPoints \gets VisitedPoints^\complement$
            \State $\{c_k\} \gets group(UnvisitedPoints, l)$ \Comment{Group unvisited points into $K$ cubes, which are in the non-empty nodes of the octree at level $l$}
            \For {$c$ in $\{c_k\}$} \Comment{Parallel for all $k$}
                \State\textit{Points} $\gets \text{GetPoints}(c)$
                \State\textit{Normals} $\gets \text{GetNormals}(c)$
                \State$P \gets mean(Points)$
                \State$r \gets \max(\|Points - P\|)$
                \State$\vec{o} = P - c.A$ \Comment{$c.A$ is the anchor of the cube $c$, shown in Fig.~\ref{fig:geometry_attributes}.}
                \State$\vec{n} \gets mean(Normals)$
                \If {$f(Points, \vec{o}, \vec{n}, r) $ \textbf{or} $l$ is the last level}
                    \State $T \gets T \cup \{(\vec{o}, \vec{n}, r)\}$
                    \State $UnvisitedPoints \gets UnvisitedPoints \backslash Points $
                \EndIf
            \EndFor
        \EndFor
        \end{algorithmic}
    \end{algorithm}
    
Given a point cloud of adequate quality (as provided in various datasets~\cite{dataset8i,gautier2023uvg}), we can build a surfel octree from the point cloud efficiently using the proposed algorithm detailed in Algorithm~\ref{alg:build}. The algorithm takes a point cloud $\{x_i = (x, y, z, r, g, b, n_x, n_y, n_z)\}$, a set of octree levels $\{l_i\}$, and a cube-splitting decision function $f$ as input. We assume the point cloud has a normal vector $\vec{n} = (n_x, n_y, n_z)$  and a color $(r, g, b)$ associated with each point. For point clouds without surface normals, we can estimate the normals using a point cloud normal estimation method~\cite{hoppe1992surface}. Our tree-construction algorithm will output a surfel octree $T = \{(\vec{o}_k, \vec{n}_k, r_k)_i\}$ defined on the designated octree levels $\{l_i\}$. In practice for a point cloud with 10-bit coordinate precision (\textit{i.e.} fully represented by a 10-level octree), we set the possible octree levels with leaf nodes to be $\{6, 7, 8\}$. That is $l_{\min}=6, l_{\max=8}$. The decision function $f: (\{x_j\}, (\vec{o}, \vec{n}, r)) \rightarrow \{True, False\}$ takes a set of points $\{x_j\}$ within a cube and an approximating surfel geometry $(\vec{o}, \vec{n}, r)$ as input, and returns a decision on whether the cube needs to be further split. In our current implementation, we  sample grid points on the surfel and compare the D1-PSNR between the grid points and the point set $\{x_j\}$ in the cube to a designated threshold $\tau$ dB. $\tau \in \{60, 62, 64, 66\}$ is used in our experiments.

\begin{figure}[t]
    \centering
    \begin{subfigure}{0.45\linewidth}\centering
        \includegraphics[width=0.65\linewidth]{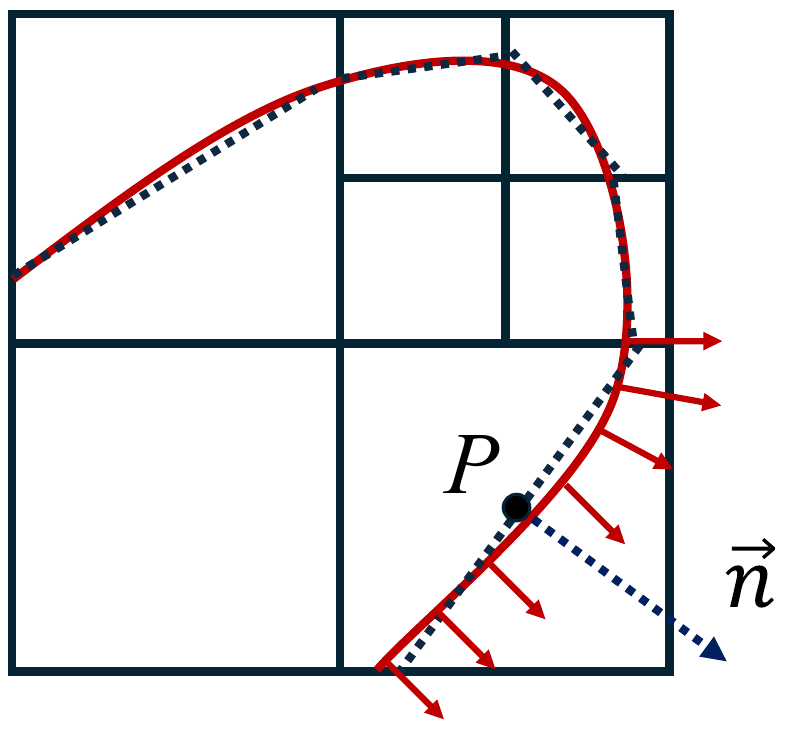}
        \caption{Average surface normal.}\label{fig:mean_normal}
    \end{subfigure}
    \begin{subfigure}{0.42\linewidth}\centering
        \includegraphics[width=0.6\linewidth]{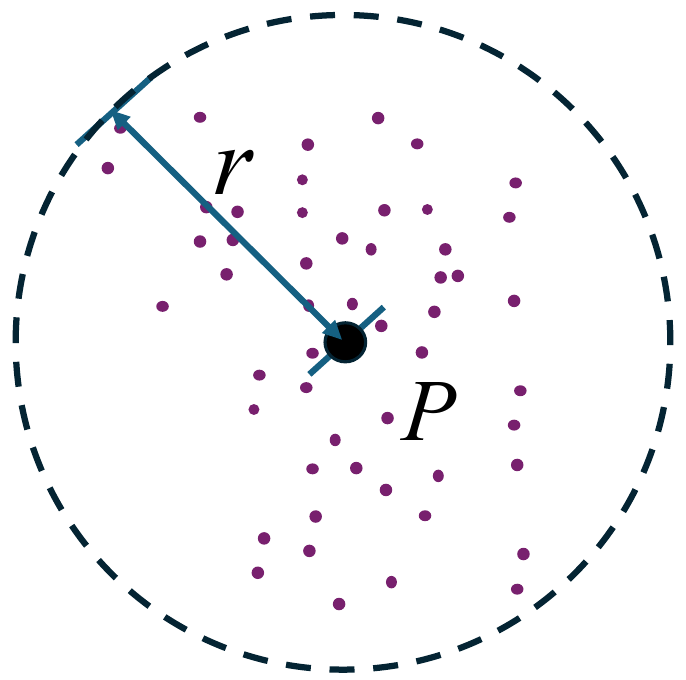}
        \vspace{3mm}
        \caption{Calculate patch radius.}\label{fig:calculate_r}
    \end{subfigure}
    \caption{Derivation of $\vec{n}$ and $r$ for the surfel. (a) The average surface normal is calculated from the normals of all points in the cube. (b) The radius $r$ is calculated as the maximum distance from the center of the surfel (namely $P$) to all points in the cube.}
    \label{fig:normal_radius}
\end{figure}

The algorithm first sorts the octree levels in ascending order. For each level $l$, the algorithm identifies non-empty cubes at that level with $K$ indicating the total number of such cubes. The algorithm also keeps track of the visited points, and remove the points in each newly identified leaf cube from the set of points. For each cube that has not been visited, the algorithm computes the mean coordinate $P$ of the point positions as the surfel center position. It then calculates the normal vector $\vec{n}$ of the surfel, by taking the average of the normals of all points in the cube, as shown in Fig.~\ref{fig:mean_normal}. It also calculates a radius $r$ for the surfel, which is the maximum distance from $P$ to all points in the cube, as shown in Fig.~\ref{fig:calculate_r}. The algorithm then checks if the surfel geometry can approximate the point cloud geometry with enough accuracy by evaluating the decision function $f$. If the decision is not to split, this surfel is added to the surfel tree and the corresponding cube is marked as a leaf cube in the octree. Our algorithm is designed to be efficient on GPU with support of data parallelism. We implement the algorithm using CUDA. Our tests on a NVIDIA RTX 4080 GPU show that the algorithm can build a surfel tree from a point cloud with 1 million points in about 0.3 seconds.

\subsection{Surfel Octree Texture}\label{sec:texture}

\begin{figure*}[t]
    \centering
    \begin{subfigure}{0.25\linewidth}\centering
        \includegraphics[width=\linewidth]{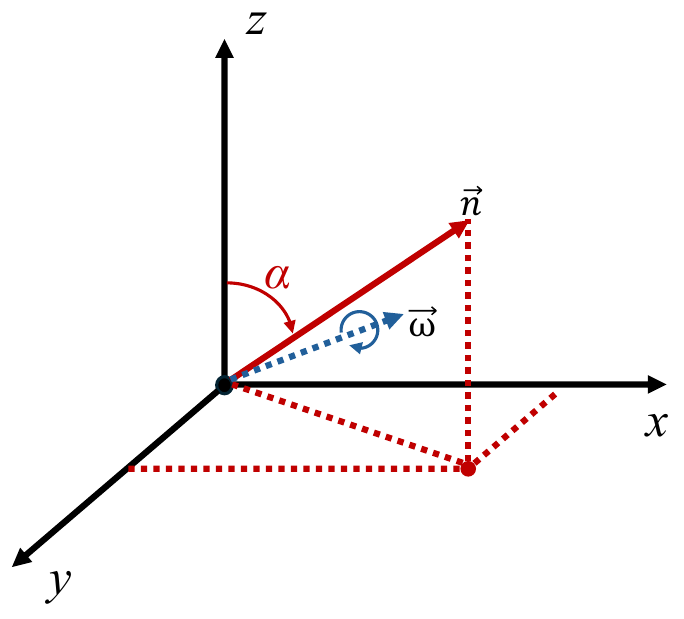}
        \caption{Canonical system.}
    \end{subfigure}
    \begin{subfigure}{0.27\linewidth}\centering
        \includegraphics[width=\linewidth]{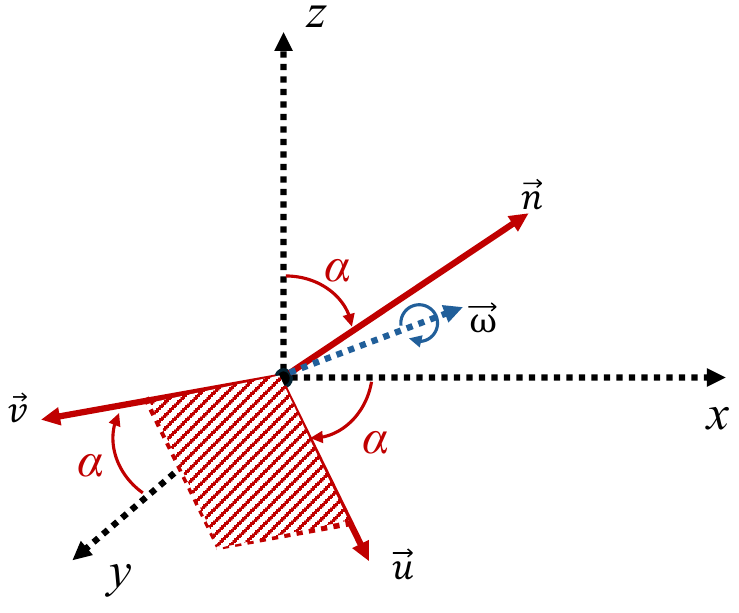}
        \caption{Surfel system.}
    \end{subfigure}
    \begin{subfigure}{0.41\linewidth}\centering
        \includegraphics[width=\linewidth]{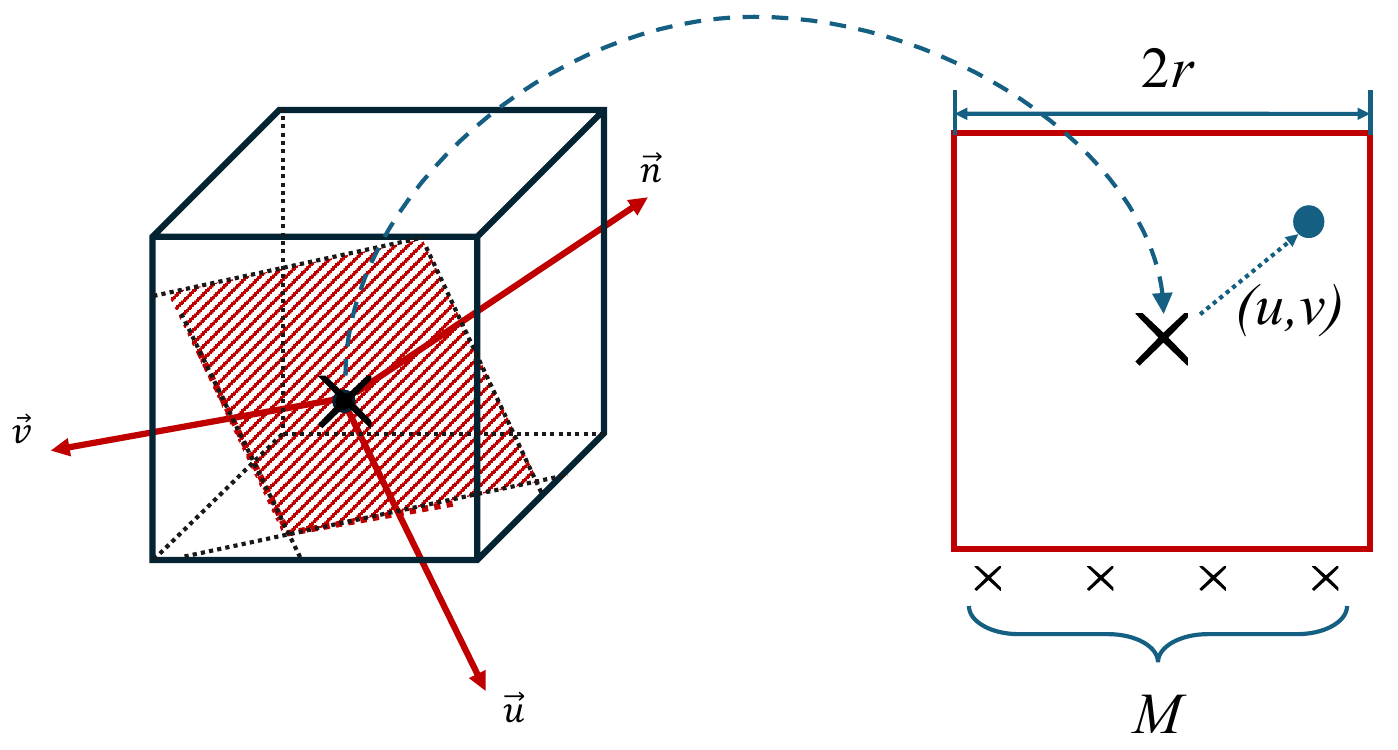}
        \caption{Texture patch mapping.}
    \end{subfigure}
    \caption{Illustration of the tangent space coordinate system and the texture representation on the surfels.}
    \label{fig:texture}
    % \vspace{-2mm}
\end{figure*}

Based on the surfel tree geometry, we further specify texture information for each surfel for photo-realistic rendering. We design the texture representation different from the two existing approaches for mesh texture, \textit{i.e.} global texture mapping and colors on vertices only, motivated by the need of flexibility and efficiency. Unlike a global texture map, the texture on the surfel tree is specified on each of the surfel, making it possible to save memory and bandwidth when the rendering does not require the information of the whole scene (\textit{e.g.} when the scene is partially occluded or is outside of the FoV). In contrast to vertex coloring, we specify a color patch on each  surfel. Hence, when the geometry specification stops at a coarse level, we can still render high frequency texture.

We first find the surfel tangent plane coordinate system, defined by three axes $(\vec{u}, \vec{v}, \vec{n})$. To minimize the burden of specifying the coordinate system as three orthogonal vectors $(\vec{u}, \vec{v}, \vec{n})$, We derive  $(\vec{u}, \vec{v})$ from $\vec{n}$ and z-axis of the world-space coordinate system. As shown in Fig.~\ref{fig:texture}, $\vec{n}$ can be considered as a unit-norm vector rotated from the z-axis $\vec{z}$ in the canonical coordinate system with a single rotation angle $\alpha = \arccos(\vec{n} \cdot \vec{z})$.
Given the normal vector $\vec{n}$, we designate a rotation axis $\vec{\omega}=\vec{n} \times \vec{z}\doteq(a, b, c)$ to be the cross product of $\vec{n}$ and the z-axis. The rotation axis $\vec{\omega}$ and the angle $\alpha$ can be used to determine the rotation matrix $\mathbf{R}$ that rotates the z-axis to $\vec{n}$, which can be specified by a quaternion $\mathbf{q}$,
\begin{equation}
    \centering
    \begin{split}
        \mathbf{q} &= \left[  \cos \frac{\alpha}{2}, a\sin \frac{\alpha}{2}, b\sin \frac{\alpha}{2}, c\sin \frac{\alpha}{2} \right].
    \end{split}
\end{equation}
By rotating the canonical coordinate system $(\vec{x}, \vec{y}, \vec{z})$ with $\mathbf{R}$ to derive $(\vec{u}, \vec{v}, \vec{n})$, and translating the origin to the center $P$ of the surfel  mean position of all points in the cube, we can find the surfel coordinate system $\mathbf{S} = \{ \vec{u}, \vec{v}, \vec{n}, P \}$, where $\vec{u}$ and $\vec{v}$ are the two orthogonal vectors in the tangent plane of the surfel.

\begin{figure}[t]
    \centering
    \includegraphics[width=0.95\linewidth]{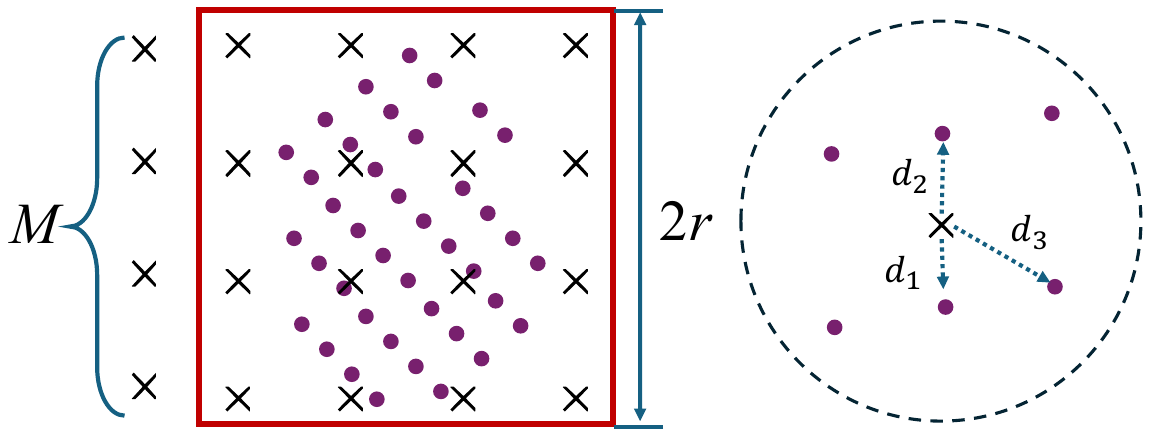}
    \caption{Represent texture using $M\times M$ pixel grid on the tangent plane.}
    \label{fig:get_texture}
\end{figure}

After defining the surfel 2D coordinate system, we can specify a texture patch using a $M\times M$ pixel grid on the tangent plane covering an area of $2r \times 2r$ as shown in Fig.~\ref{fig:get_texture}. In our experiments, we use $M = (12, 8, 4)$ for $l = (6, 7, 8)$, respectively. We obtain the color value of each pixel by interpolating a color at the pixel coordinate from $K$ nearest neighbors in the point cloud, as illustrated in Fig.~\ref{fig:get_texture}. To avoid discontinuity between adjacent surfels, we find the $K$ nearest neighbors among points in the surfel's own cube and neighboring cubes where the surfel plane can potentially extend to. We set a threshold for the distance of a point to the surfel plane and ignore the points that are too far away. The color of each pixel in the texture patch is then calculated as the weighted average of the colors of the $K$ nearest neighbors, where the weights are inversely proportional to the distances to the points. We use $K=3$ in the experiments.

\section{Efficient Rendering}\label{sec:rasterization}

The proposed TeSO can be rendered from any view point using efficient rasterization. We first sort all potential intersecting surfels with a camera ray for a pixel in its z-buffer (ordered in the distance between the camera center and surfel center). The intersection between a ray with origin $\vec{O}$ and direction $\vec{d}$ and a surfel with center point $\vec{P}$ and normal vector $\vec{n}$ can be computed by
\begin{equation}
    \begin{split}
        \vec{v} = \vec{P} - \vec{O}; \;
        t = \frac{\vec{n} \cdot \vec{v}}{\vec{n} \cdot \vec{d}};\;
        \vec{P}_{\text{hit}} = \vec{O} + t \vec{d}
    \end{split}
\end{equation}
We check whether the hit point $\vec{P}_{\text{hit}}$ is inside the bounding cube of the surfel and if the intersecting point $P_{\text{hit}}$ is within radius $r$ to the surfel center $P$. If it is, we shade the rendered pixel by the bilinearly interpolated color from four nearest pixels to the intersecting point in the texture patch  (discussed in Section~\ref{sec:texture}), and we stop checking following possible intersections in the z-buffer for this pixel. If the hit point is within the soft area defined in Section~\ref{sec:geometry}, we multiply it with the weight $\alpha$ and continue to the next intersected surfel in the z-buffer. If the ray does not hit a surfel, we also continue to check the next surfel in the z-buffer, until all possible intersecting surfels are checked. If a pixel is not covered by any surfel, we assign it to the predefined background color.

\section{Coding of TeSO}\label{sec:compression}

\begin{figure}[t]
    \centering
    \includegraphics[width=0.99\linewidth]{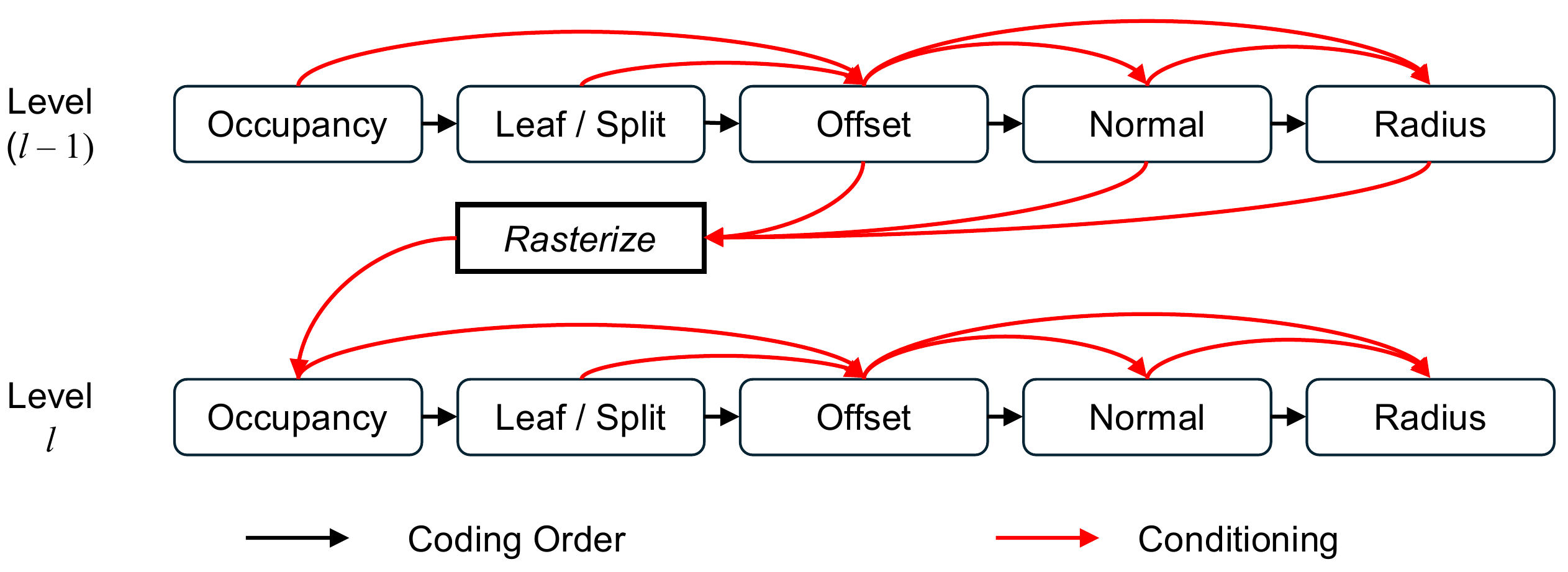}
    \caption{Coding order and conditioning dependency of geometry attributes.}
    \label{fig:coding_dependency}
\end{figure}

In this section, we present the techniques to compress the proposed TeSO representation. We assume the sender has the original colored point cloud, from which the sender can build surfel trees with different geometry granularity parameters, and then encode the texture with different quantization parameters, to meet the requirement of the target bit-rate. To control the bit-rate of the geometry, the sender has the option to choose the following parameters: (1) the octree levels $l_{\min}$ and $l_{\max}$, which determine the granularity of the geometry; (2) the D1-PSNR threshold $\tau$ for the decision function $f$, which also controls the granularity of the geometry. 

The senders build the surfel octree with the designated parameters, and then quantize the geometry attributes.
For the offset $\vec{o}$ comprised of $(\delta_x, \delta_y, \delta z) \in [0, b)^3$, where $b$ is the width of the octree cube at a particular level, we conduct a scalar quantization for each of the $\delta$ to a quantization step of $0.5$, effectively quantizing each values to $2b$ possible values.
% The values of $(\vec{o}, \vec{n}, r)$ are quantized to $(5, 4, 8)$ bits, respectively. These bit sizes and corresponding quantization steps are chosen such that the it does not affect rendering quality compared to using the original attributes. For the offset $\vec{o}$ in an octree cube of width $b$, we conduct a scalar quantization with quantization step $b / 32$.
% , in effect quantizing each values to 32 possible values in the range of $[0, b-1]$, which is then normalized to $[0, 31]$ for all levels in the octree.
For the normal vector, we employ the octahedral quantization~\cite{meyer2010floating} to represent each normal vector $\vec{n} = (n_x, n_y, n_z)$ to $(u, v)$, where,
\begin{equation}
\begin{split}
    &p_x = \frac{n_x}{|n_x| + |n_y| + |n_z|}, \text{ } p_y =  \frac{n_y}{|n_x| + |n_y| + |n_z|},
\end{split}    
\end{equation}
\begin{equation}
u = \begin{cases}
        p_x & \text{if } n_z \geq 0 \\
        (1 - p_y) \operatorname{Sgn} (p_x) & \text{if } n_z < 0
    \end{cases},\text{ }
v = \begin{cases}
        p_y & \text{if } n_z \geq 0 \\
        (1 - p_x) \operatorname{Sgn} (p_y) & \text{if } n_z < 0
    \end{cases}.
\end{equation}
We then use scalar quantization on $(u, v), u, v \in [-1, 1]$, with quantization step $1/64$. For the radius $r \in \left(0, \frac{\sqrt{3}}{2}b\right)$, we quantize the scalar value with quantization step $1/16$. These bit sizes and corresponding quantization steps are chosen such that the it does not affect rendering quality compared to using the original attributes.

Since the  surfel texture patches depend on the  surfel octree geometry, in the compression process, we first build the surfel tree geometry with the given geometry parameters. The resulting quantized geometry is losslessly coded using a learned entropy model. We then find the texture patches based on the quantized surfel geometry and compress them with the given color quantization parameter $Q_t$. We use $l_{\min}=6$ in the experiments. We find the best combination of $(l_{\max}, \tau, Q_t)$ that reaches the best rendering quality for a target rate (to be discussed later). In the following, we first present the proposed compression pipeline for the surfel octree geometry, and then discuss the texture compression.

\subsection{Geometry Compression}

\begin{figure}
\centering
\includegraphics[width=1\linewidth]{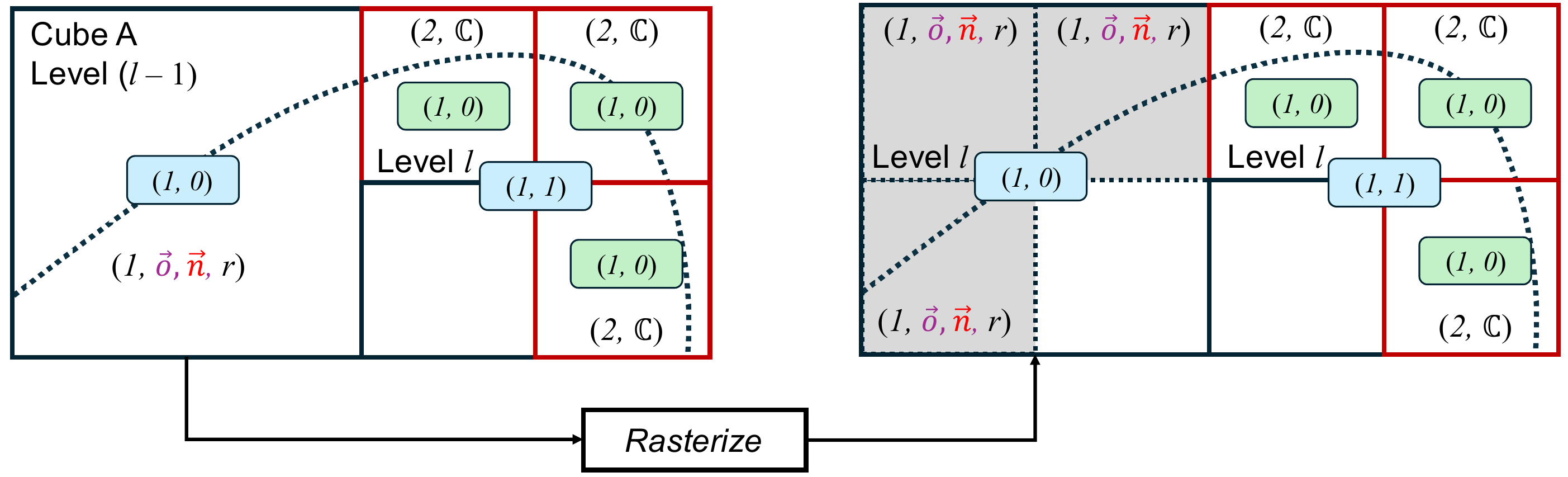}
\caption{Illustration of context preparation involving neighboring cubes that stop at the parent (\textit{i.e.} $(l-1)$) of the current coding level $l$.}
\label{fig:context_rasterize}
\end{figure}

\begin{figure}[t]
    \centering
    \centering
    \includegraphics[width=1\linewidth]{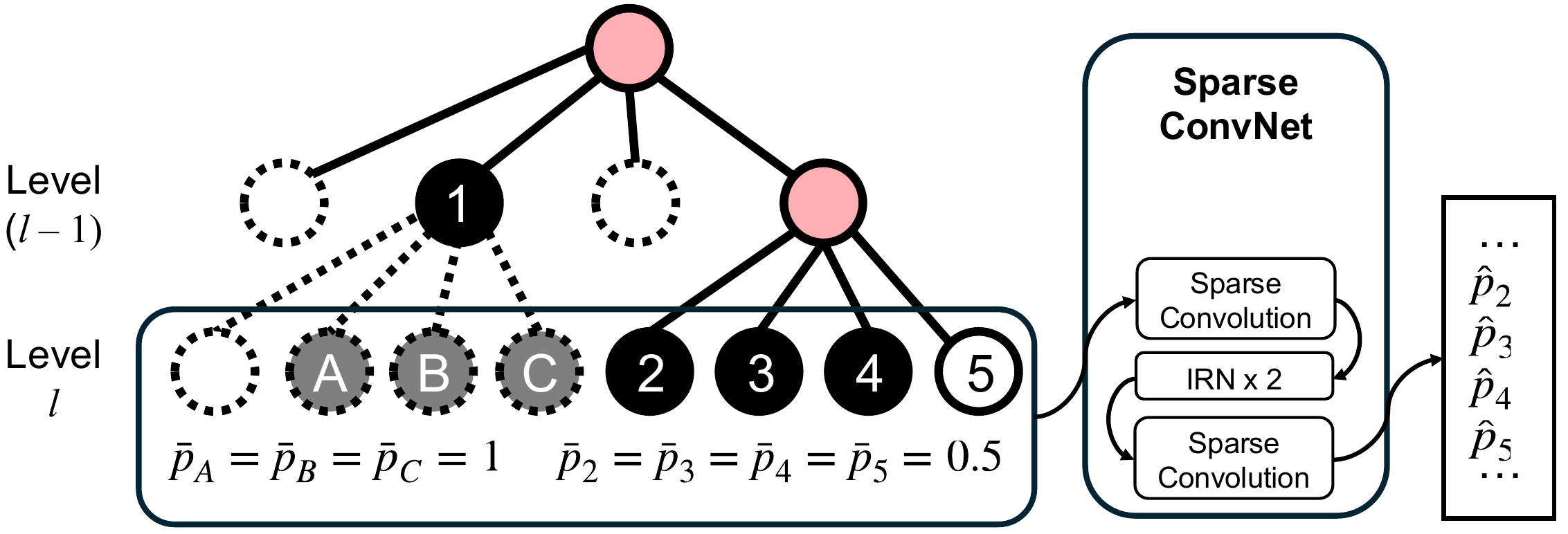}
    \caption{Architecture of the neural network for coding the occupancy.}\label{fig:occupancy_net}
\end{figure}

To losslessly compress the surfel octree structure with attributes specified in Fig.~\ref{fig:design}, we need to encode the occupancy bit, the leaf node flag, and the surfel attributes $(\vec{o}, \vec{n}, r)$.
Let $l_{\min}$ be the lowest coarse level of the octree. We define the base octree to be the subtree from the root node to level $l_{\min}$. Note that the base octree is similar to an octree used to describe a point cloud, where every node in the intermediate levels is either an empty node or a non-empty node (to be further split). Only in the final level ($l_{\min}$), a non-empty node needs to be labeled using a leaf node flag (with ``0" indicating a leaf node and ``1" indicating a node to be further split).  Therefore, we can use G-PCC~\cite{tmc13} to code the occupancy for the base octree. We will discuss how to code the leaf flag bit-stream and the surfel geometry attributes for leaf nodes below.

For the geometry attributes (including occupancy, leaf status, and surfel geometry attributes for leaf nodes) of the remaining levels, we employ conditional entropy estimation and arithmetic coding. The geometry attributes at each level of the octree is coded conditioned on its parent level, and within each level we employ conditional coding across different geometry attributes types. The coding dependency and conditioning chain is shown in Fig.~\ref{fig:coding_dependency}. 

Suppose we are coding the geometry information at level $l (l > l_{\min})$ given information of level $(l-1)$. We first encode the occupancy bits. We prepare the context for the occupancy bits by rasterizing the octree nodes at level $(l-1)$ to the current level $l$. We show an example of the rasterization process in Fig.~\ref{fig:context_rasterize}. \textit{Cube A} is a node that stops at level $(l-1)$, which has the occupancy bit ``$1$" and the splitting bit ``$0$". To extend it as a context for convolutions at level $l$, we first check if each of the sub-cube intersect with the surfel and assign those that intersect (marked gray in Fig.~\ref{fig:context_rasterize}) an occupancy probability of 1, as shown in Node A, B, and C in Fig.~\ref{fig:occupancy_net}. For the unknown nodes (split from level $l-1$, shown as Node 2, 3, 4, and 5 in Fig.~\ref{fig:occupancy_net}), we assign them initial probability of $0.5$. We employ a 3D sparse convolutional neural network to predict the occupancy probability for unknown nodes, with these probabilities as input attributes. The final occupancy is coded using an arithmetic encoder with the predicted probability, \textit{i.e.} $\hat{p}_2, \hat{p}_3, \hat{p}_4, \hat{p}_5$.

Once the occupancy bits are coded and known, we encode the leaf node flag for each occupied node (not including the contextual nodes). We first make the leaf node flag into a sequence of bits through Morton order traversal of the occupied nodes. We then use Context Adaptive Binary Arithmetic Coding (CABAC)~\cite{schwarz2021cabac} to code the bit sequence. This method is used to code the leaf flag stream of level $l_{\min} \leq l < l_{\max}$.

\begin{figure*}[t]
    \centering
        \centering
        \includegraphics[width=0.9\linewidth]{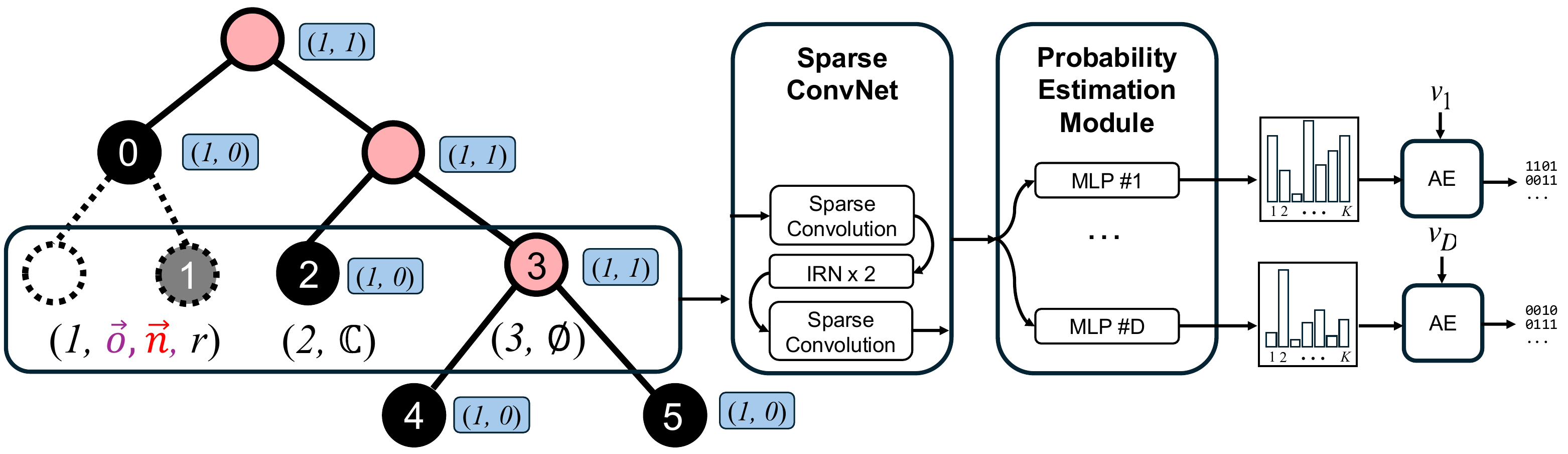}
    \caption{The compression pipeline of the surfel octree geometry attributes with the rasterized context.}\label{fig:paradigm}
\end{figure*}

Once the occupancy and leaf node flags are coded, we  proceed to code the geometry attributes for the leaf nodes with the following steps:
\begin{enumerate}
    \item \textbf{Lower Level Context Preparation}. Similar to the context preparation for occupancy coding, for nodes stopped at a level lower than $l$, we find the \textit{virtual} children nodes at level $l$ that carry the contextual information, as illustrated in Fig.~\ref{fig:context_rasterize}. Each of these nodes is attached with the already coded geometry attributes $(1, \vec{o}, \vec{n}, r)$ of the ancestor node, with ``1'' indicating a virtual node.
    
    \item \textbf{Current Level Context Preparation}. As shown in Fig.~\ref{fig:coding_dependency}, for each leaf node at level $l$, we first encode the offsets $\{ \vec{o}_i\}$, then the surface normal vectors $\{ \vec{n}_i\} $, and finally the radius $\{r_i\}$. The coding of the offsets is conditioned on the occupancy bits and the leaf node flags. So the input $\mathbb{C} = \{2, \mathbf{0}, \mathbf{0}, \mathbf{0}\}$ in Fig.~\ref{fig:paradigm}, where ``2" indicates it is a leaf node. The coding of the surface normal vectors is conditioned on the offsets, so $\mathbb{C} = \{2, \vec{o}, \mathbf{0}, \mathbf{0}\}$. And the coding of the radius is conditioned on both the offsets and the surface normal vectors, so $\mathbb{C} = \{2, \vec{o}, \vec{n}, \mathbf{0}\}$. 

    \item \textbf{Probability Estimation}. We use a sparse convolutional neural network (Fig.~\ref{fig:paradigm}) acting on all non-empty nodes to estimate the probability distribution of each attribute to be coded. The input to the network includes three types of nodes at the coding level $l$: (i) virtual nodes from lower level with known geometry attributes (marked \#1 in Fig.~\ref{fig:paradigm}), which have full attributes; (ii) the leaf nodes to be coded (marked \#2), which have context $\mathbb{C}$; (iii) nodes that will be further split into higher levels (marked \#3), which have no known attributes and are assigned an all zero features. We then conduct sparse convolutions on these nodes with assigned context values as their initial features, and use the final features at each leaf node to predict the categorical distribution of each quantized attribute parameter. We use $D$ classifiers (each a multi-layer perceptron or MLP), where $D$ is the number of parameters for the attribute to be coded (\textit{i.e.} $D=(3, 2, 1)$ for $(\vec{o}, \vec{n}, r$), respectively), and each classifier outputs $K_q$ probabilities, where $K_q$ indicates the number of possible quantized values for each parameter. 
    \item \textbf{Entropy Coding}. With the categorical distribution predicted by the probability estimation module, we use an arithmetric encoder (AE) to losslessly encode the quantized attributes into bit-streams.
\end{enumerate}

The achievable lower bound of the entropy coding is given by the cross entropy between the categorical distribution and the one-hot distribution of the quantized attributes, as,
\begin{equation}
    \centering
    \begin{split}
        \mathcal{L_{\text{entropy}}} & = \sum_{i=1}^3 H(\hat{o}_i) + \sum_{i=1}^2 H(\hat{n}_i) + H(\hat{r})\\
        H(x) &= -\sum_{i=1}^{N} \sum_{j=1}^{K_{q,x}} \mathds{1}_{x_i=j} \log_2\left(p(x_i=j|\mathbb{C})\right), \\
    \end{split}
\end{equation}
where $N$ is the number of nodes at the current level, $K_{q,x}$ is the number of possible quantized values for the attribute $x$, and $x_i$ is the quantized value of a specific attribute entry for node $i$. We train the neural networks to minimize this cross entropy loss, which minimizes the final bit-rate for lossless compression.

\subsection{Training}
The only component in the proposed scheme that requires training is the entropy model (\textit{i.e.} the neural networks that predict occupancy probabilities and the categorical probability distribution for surfel geometry attribute coding). We leverage the UVG-VPC dataset~\cite{gautier2023uvg} to build the training TeSOs. This dataset provides 12 colored point cloud sequences, each with 250 frames. We remove invalid frames (\textit{i.e.} 44 out of 3000 frames where point clouds are not appropriately built due to capturing errors) by manually checking the rendering results. We set aside 1\% of the training set as validation set. We build the surfel octree datasets using a range of values of $\tau \in \{60, 62, 64, 66\}$. We train separate entropy models for each of the octree levels $\{6, 7, 8\}$, and for each of the geometry attributes $(\vec{o}, \vec{n}, r)$, respectively.

\subsection{Texture Compression}

\begin{figure}
    \centering
    \begin{subfigure}{0.43\linewidth}
        \centering
        \includegraphics[width=1\linewidth]{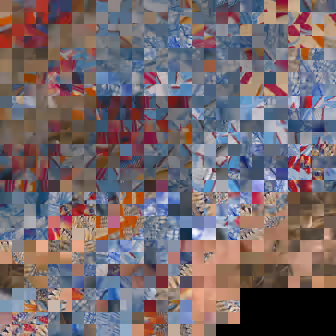}
        \caption{$l = 6$}
    \end{subfigure}
    \begin{subfigure}{0.45\linewidth}
        \centering
        \includegraphics[width=1\linewidth]{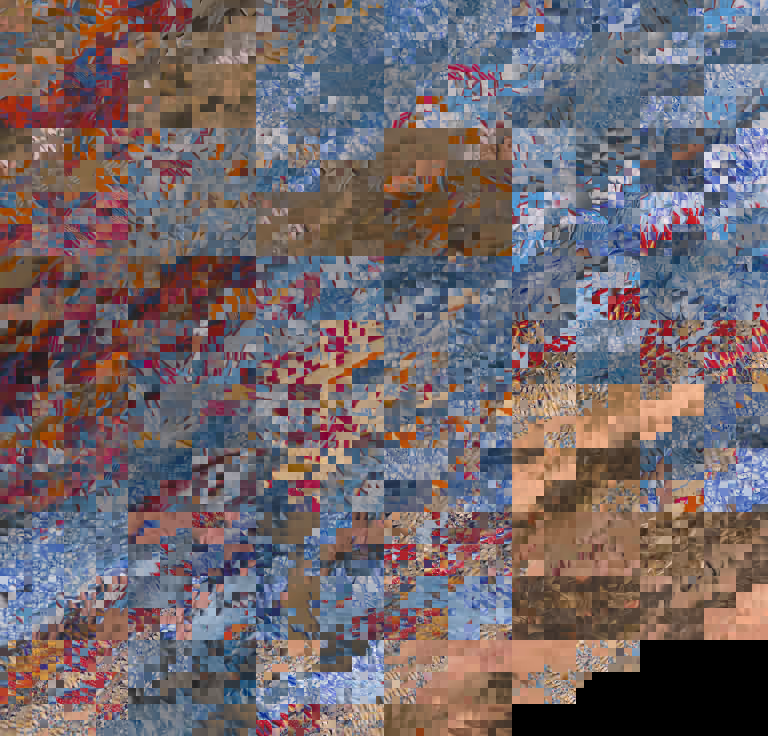}
        \caption{$l = 7$}
    \end{subfigure}
    \caption{Packed texture patches at two different levels.}
    \label{fig:packed}
\end{figure}

Once the surfel octree geometry is determined, the texture patches on the surfels can be calculated based on the quantized geometry and encoded. We adopt two standard approaches to encode the texture patches: 
\begin{enumerate}
    \item \textbf{Standard Video Codec}. We first pack the surfel texture patches into an image using the Morton order following the approach of~\cite{lee2020groot}. This packing method will organize patches that are close in the 3D space into blocks, making it more suitable for standard hybrid codec to compress them as images. An example of the packed images at two different levels is shown in Fig.~\ref{fig:packed}. We use the intra coding tool of the AV1 video codec~\cite{han2021technical} to encode the images in YUV 4:4:4. Since the Morton order is only dependent on the geometry, the decoded packed images can be inversely reconstructed into surfel texture patches.
    \item \textbf{Coding as Colored Point Cloud}. Given the surfel octree geometry, each pixel in the texture patch can be considered as a colored point in the 3D space. Therefore, the entire TeSO can be rasterized into a colored point cloud. We can then use the G-PCC~\cite{tmc13} to encode the colored point cloud. Since the geometry is already encoded and made available to both the encoder and decoder, we only need to transmit the bits for the color attributes in the point cloud bit-stream.
\end{enumerate}

We analyze the rate-distortion performance and decoding complexity of the two approaches in Section~\ref{sec:result}. We found that the G-PCC-based approach generally achieves a better R-D performance and better visual quality than the standard video codec, but it takes longer time to decode.

\section{Experimental Results}
\label{sec:result}

\subsection{Experimental Settings}

\subsubsection{Rendering Quality Evaluation}

\begin{figure}
    \centering
    \includegraphics[width=0.7\linewidth]{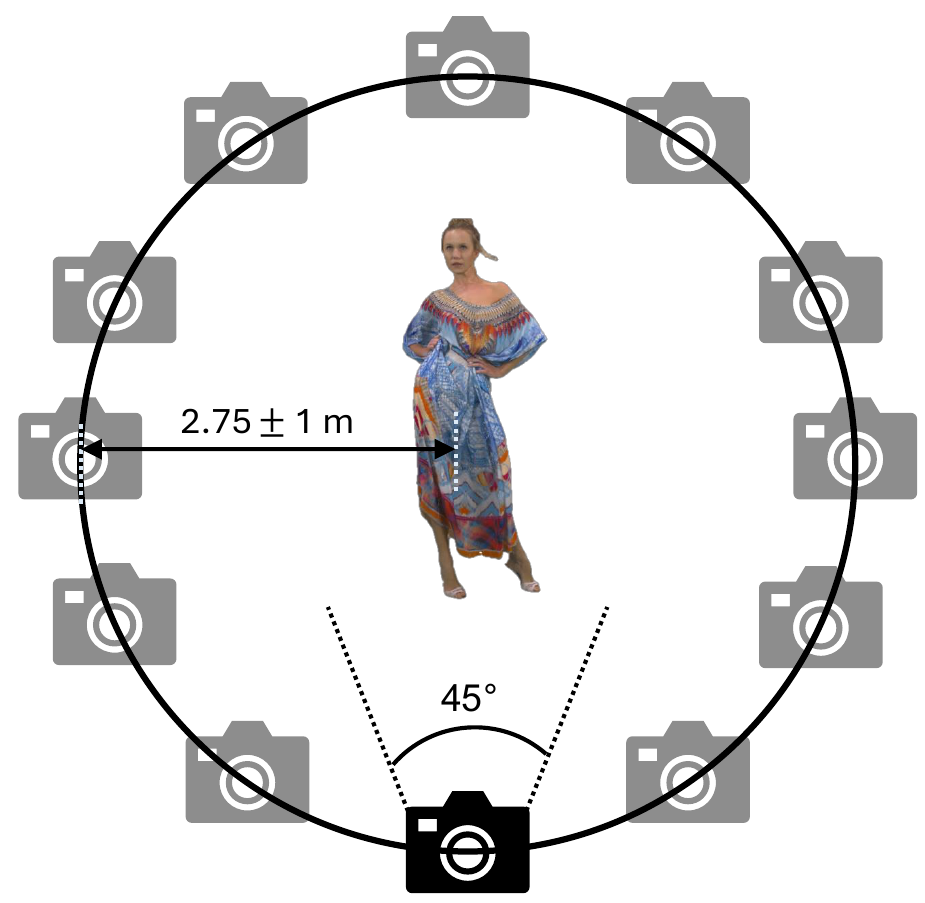}
    \caption{Rendering camera trajectory setup.}
    \label{fig:rendering}
\end{figure}

Since the proposed scheme does not reconstruct point clouds but directly decodes to surfel primitives for rendering, we evaluate the rendering quality as the distortion metric. We use the rendering setup shown in Fig.~\ref{fig:rendering}. The camera is initially placed on a circle with a radius of 2.75m. We set the camera Field of View (FoV) to 45 degrees, which guarantees that the figure is fully captured and takes up the entire screen. We designate a circular trajectory to the camera, which always faces the subject in the scene. We simulate a real VR usecase by perturbing the distance from the camera to the center of the circle by $\pm 1$m, and generate the camera trajectory.  We use the same camera trajectory for all rendering. To evaluate the sensitivity of different methods to the rendering resolution, we render each scene to two different resolutions: $1024\times 1024$ and $1920\times 1920$.

We leverage the 8iVFB~\cite{dataset8i} dataset for evaluation, which contains 10-bit color point clouds of 4 subjects. We first use Poisson surface reconstruction~\cite{kazhdan2013screened} to generate the Poisson Mesh from the original point cloud, and then rasterize the meshes to generate the ground truth views following the camera setup. Since small differences in the actual location of the boundaries between the 3D scenes and the background can lead to large differences to pixel-based metrics like PSNR and SSIM, in this work, we use LPIPS~\cite{zhang2018unreasonable} between the rendered views and the ground truth views to evaluate the rendering quality, since it is closer to human perception and less sensitive to pixel-level shifting.

\subsubsection{Baseline Methods}

We compare the proposed method with G-PCC~\cite{tmc13} (version 22.0) and the joint point cloud compression and rendering method B2P~\cite{hu2024bits}. For G-PCC, we use the lossy-geometry-lossy-color configuration, where we evaluate a wide range of downsampling ratio for the point cloud (controlling geometry accuracy) and quantization parameters~(QP) (controlling color accuracy) to obtain a large set of rate-distortion (R-D) points. To render G-PCC decoded point clouds, we employ two fast rendering methods:

\begin{enumerate}
    \item \textbf{G-PCC OpenGL}. The G-PCC decoded point clouds are rendered using the standard point cloud rendering method implemented in Open3D~\cite{zhou2018open3d} and OpenGL, \textit{i.e.} with the \texttt{GL\_Points} primitive~\cite{openglprimitive}. It splats each point in a point cloud with a fixed $m \times m$ pixel square patch, where $m$ can be controlled by the user. Note that for different point cloud densities, rendering resolution, FoV and camera distance, the optimal $m$ vary greatly. We manually choose $m$  such that it reaches the best quality at the initial camera location (2.75m, front-facing) and fix it for the entire camera trajectory. % The rendering is very fast (QUANTIFY)
    \item \textbf{G-PCC P2ENet}. We employ the learned point cloud rendering method proposed in~\cite{hu2024low}, which learns a neural network to predict the 3D Gaussian parameters for each point in the point cloud for improved rendering quality. The network is optimized to render at resolution $1024\times 1024$, and also trained to handle quantization errors caused by G-PCC lossy geometry downsampling. Thus it generally produces better rendering quality than G-PCC OpenGL for the same rate point. The conversion from point cloud to the Gaussians take 0.13 second. %The rendering of each view from Gaussians takes 3 milliseconds.
\end{enumerate}

For bit-rates, we report bits-per-point (bpp) for all methods, normalized by  the number of points in the original point cloud.

\subsection{Rendering Quality Evaluation}

\begin{figure*}
\centering
    
    \begin{subfigure}{0.375\linewidth}
        \centering
        \includegraphics[width=0.95\linewidth]{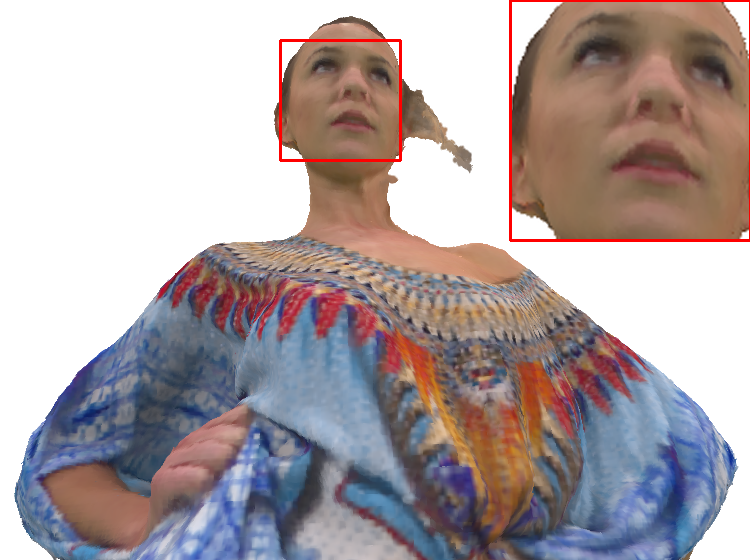}
        \caption{}
    \end{subfigure}
    \begin{subfigure}{0.375\linewidth}
        \centering
        \includegraphics[width=0.95\linewidth]{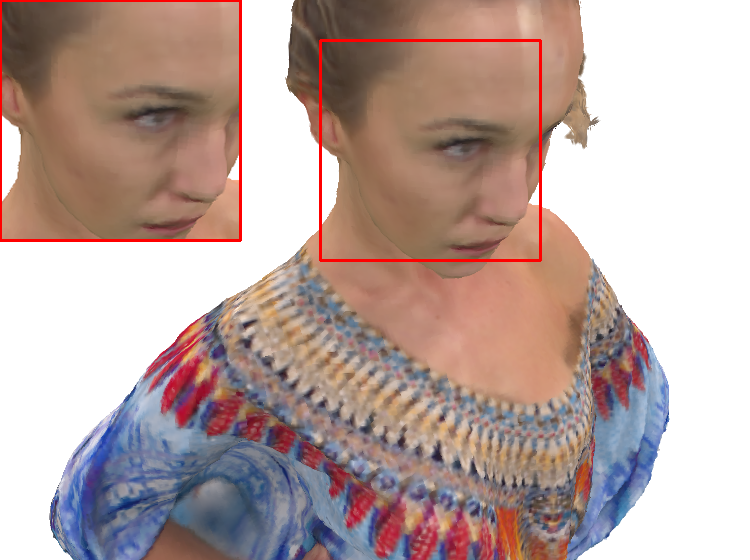}
        \caption{}
    \end{subfigure}
    \begin{subfigure}{0.235\linewidth}
        \centering
        \includegraphics[width=0.9\linewidth]{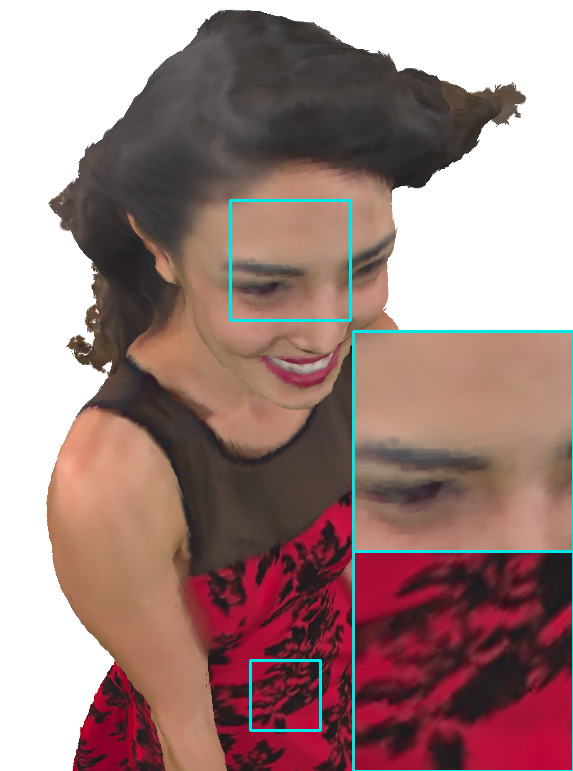}
        \caption{}
    \end{subfigure}

    \begin{subfigure}{0.375\linewidth}
        \centering
        \includegraphics[width=0.95\linewidth]{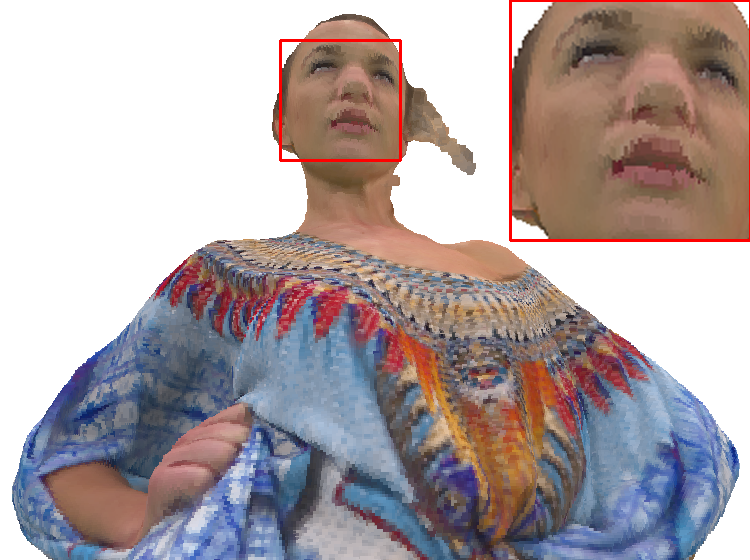}
        \caption{}
    \end{subfigure}
    \begin{subfigure}{0.375\linewidth}
        \centering
        \includegraphics[width=0.95\linewidth]{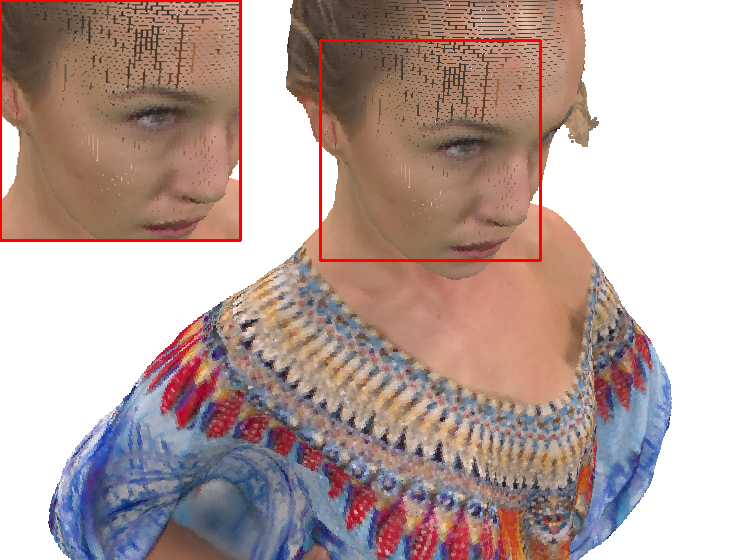}
        \caption{}
    \end{subfigure}
    \begin{subfigure}{0.235\linewidth}
        \centering
        \includegraphics[width=0.9\linewidth]{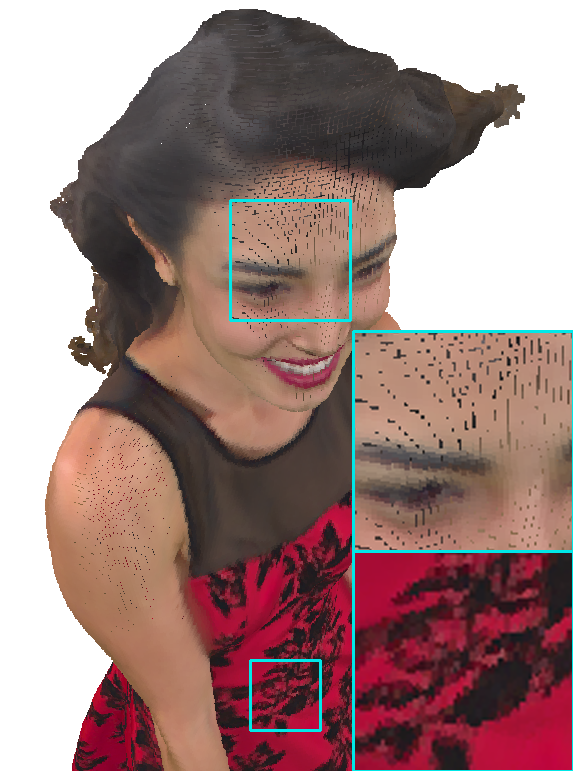}
        \caption{}
    \end{subfigure}
    
    \caption{(a), (b) and (c) are rendering results of TeSO built from the original point clouds (\textit{longdress} and \textit{redandblack} in the 8iVFB dataset). (d), (e), and (f) are rendering results of the original point clouds using OpenGL. OpenGL uses a globally set point size, which causes crowdedness and gaps in renders. Even if the point size can be manually adjusted for each view, these artefacts will still appear in the same render due to the fact that all points use the same size, as shown in (f). TeSO representation does not need adjustment to parameters during rendering.}
    \label{fig:original_render}
\end{figure*}

We first demonstrate that the proposed TeSO can render better images than the standard point cloud rendering approach OpenGL. We build TeSO from the original point cloud, with quantized surfel geometry attributes, and $\tau = 66$. We render TeSO at different camera views, and compare the renders to results rendered using OpenGL on the original point cloud. For OpenGL, we manually choose a point size $m$  at an initial view to find a balance between the gaps and crowdedness. As shown in Fig.~\ref{fig:original_render}, OpenGL suffers from visible gaps and overcrowdedness in the rendered images because it renders points as fixed-size square splats globally. These two kinds of artifacts can occur in the same view, seen from points with different distance to the camera. In contrast, the proposed TeSO  does not require manually setting any parameters, and can render correct textured surface at arbitrary camera poses and camera settings (\textit{i.e.} focal length, resolutions). This demonstrates that TeSO can serve as a better representation for rendering.

\subsection{Ablation Study on Geometry Coding}

\begin{table}[h]
\centering
\caption{Impact of different conditioning combinations (first row) on the compression ratio (compressed bits / raw bits) for the geometry attributes.}\label{tab:ablation}
\begin{tabular}{lccc}
\hline
{Coding Attributes} & {None} & {Offset} & {Offset \& Normal} \\
\hline
Offset         & 40\% & N/A & N/A \\
Normal  & 47\% & 31\% & N/A \\
Radius         & 66\% & 42\% & 37\% \\
\hline
\end{tabular}
\end{table}

As described in Section~\ref{sec:compression}, when we compress the geometry attributes (namely the normal vector $\vec{n}$ and the radius $r$), the probability estimation and entropy coding are conditioned on previously coded attributes at the same node (namely $\{\vec{o}\}$ and $\{\vec{o}, \vec{n}\}$, respectively). Table~\ref{tab:ablation} evaluates the benefits from such sequential conditional encoding. For each level of the octree, we first encode the offsets. When encoding the normal vectors, conditioning on previously coded offsets results in 16\% additional compression ratio. With the coded normal vectors and offsets, the bit-rate for the radius can be further reduced by 29\%. 
%It shows the necessity to conduct recursive coding for the geometry attributes. 

\subsection{Rate-Distortion Performance}

We evaluate the rate-distortion performance of the proposed method against G-PCC and B2P. We build the TeSOs with different D1-PSNR thresholds $\tau \in \{60, 62, 64, 66\}$. We then compress each quantized TeSO geometry losslessly using the learned entropy model. We compress the texture patches using the two different methods discussed in Section~\ref{sec:compression}, and vary the QP to obtain different R-D points. For each method and each sequence, we calculate a convex hull of the R-D points obtained with different $\tau$ and $QP$. We then generate an average R-D curve over the four sequences by averaging the rates and distortions, respectively, at a similar rate-distortion slope. 
%enumerate a range of rate-distortion Lagrange multiplier $\lambda$ and average the R-D points with the same $\lambda$ from different point clouds to obtain the final averaged R-D curve. 

\begin{figure*}[t]
    \centering
    % \begin{subfigure}{0.328\linewidth}
    %     \centering
    %     \includegraphics[width=\linewidth]{figures/rd_figures/avg_512_lpips_vs_bpp.pdf}
    % \end{subfigure}
    \begin{subfigure}{0.49\linewidth}
        \centering
        \includegraphics[width=\linewidth]{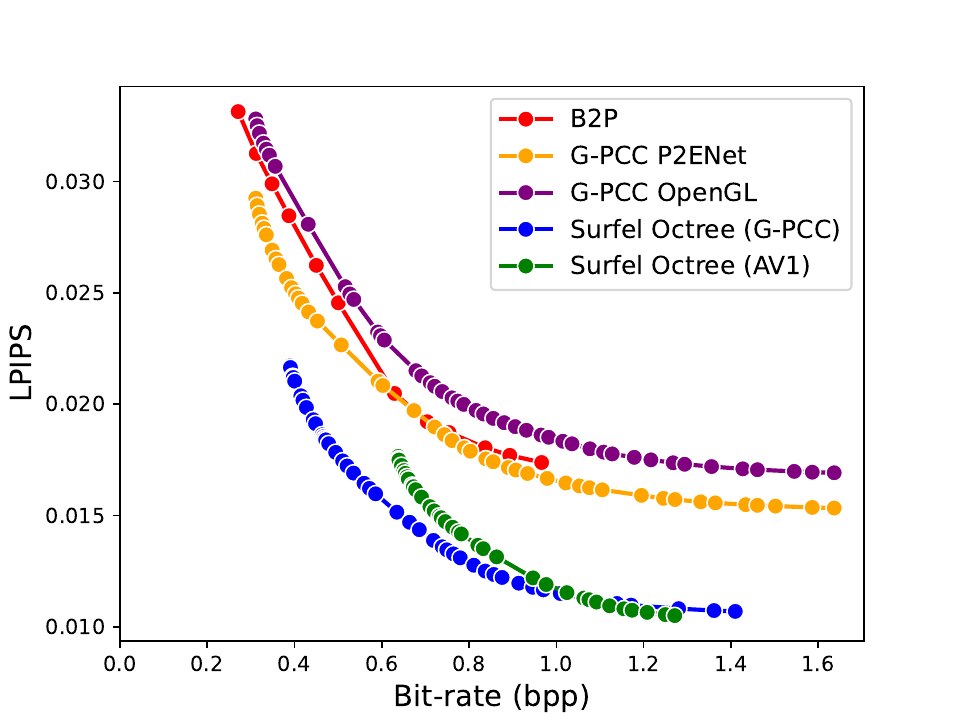}
    \end{subfigure}
    \begin{subfigure}{0.49\linewidth}
        \centering
        \includegraphics[width=\linewidth]{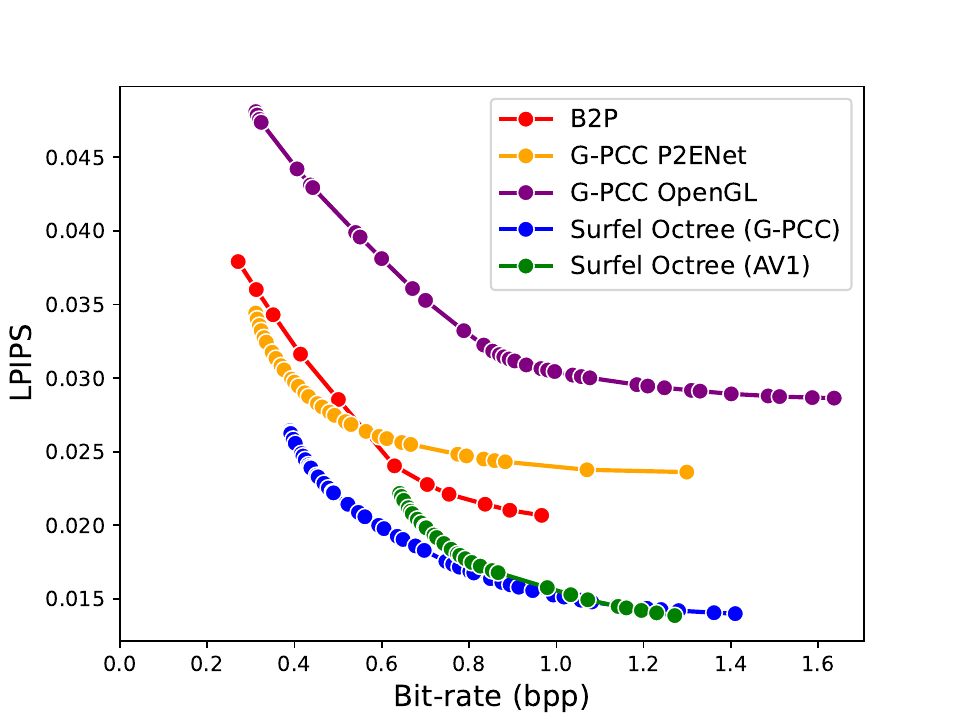}
    \end{subfigure}

    % \begin{subfigure}{0.32\linewidth}
    %     \centering
    %     \includegraphics[width=\linewidth]{figures/rd_figures/avg_512_psnr_vs_bpp.pdf}
    % \end{subfigure}
    % \begin{subfigure}{0.32\linewidth}
    %     \centering
    %     \includegraphics[width=\linewidth]{figures/rd_figures/avg_1024_psnr_vs_bpp.pdf}
    % \end{subfigure}
    % \begin{subfigure}{0.32\linewidth}
    %     \centering
    %     \includegraphics[width=\linewidth]{figures/rd_figures/avg_1920_psnr_vs_bpp.pdf}
    % \end{subfigure}

    % \begin{subfigure}{0.32\linewidth}
    %     \centering
    %     \includegraphics[width=\linewidth]{figures/rd_figures/avg_512_ssim_vs_bpp.pdf}
    % \end{subfigure}
    % \begin{subfigure}{0.32\linewidth}
    %     \centering
    %     \includegraphics[width=\linewidth]{figures/rd_figures/avg_1024_ssim_vs_bpp.pdf}
    % \end{subfigure}
    % \begin{subfigure}{0.32\linewidth}
    %     \centering
    %     \includegraphics[width=\linewidth]{figures/rd_figures/avg_1920_ssim_vs_bpp.pdf}
    % \end{subfigure}

    % \begin{subfigure}{0.328\linewidth}
    %     \centering
    %     $512\times 512$
    % \end{subfigure}
    \begin{subfigure}{0.49\linewidth}
        \centering
        $1024\times 1024$
    \end{subfigure}
    \begin{subfigure}{0.49\linewidth}
        \centering
        $1920\times 1920$
    \end{subfigure}

    \caption{Rate-distortion performance of the proposed method against G-PCC and B2P, under different resolutions. Labels in the parentheses after ``Surfel Octree" denote the two texture coding schemes. Vertical axis is the rendering distortion, the average LPIPS among all rendered views.}\label{fig:rd}
\end{figure*}

As shown in Fig.~\ref{fig:rd}, the proposed TeSO representation combined with the compression methods achieves substantially better tradeoff between the bit-rate and rendering distortion compared to G-PCC with OpenGL for rendering under both rendering resolutions. G-PCC with the learned 3D Gaussian-based renderer (P2ENet) boosts the rendering quality significantly compared to the standard OpenGL renderer, but it is still not as good as the proposed method, due to the fact that the lossy geometry of G-PCC reduces the number of points in the point cloud, which leads to a loss of details in the rendering. The proposed TeSO compression also outperforms the end-to-end learned B2P in R-D performance, thanks to the capability of representing dense texture.

Fig.~\ref{fig:rd} also shows the comparison between two different surfel texture coding methods. Generally the G-PCC-based texture coding method achieves better performance than coding texture patches with the standard video codec. This is because G-PCC is able to better exploit spatial correlations among colors distributed in 3D. Note that AV1-based texture decoding is much faster than G-PCC-based texture decoding, as shown in Table~\ref{tab:decoding_time}. We also show that the proposed method achieves a comparable decoding latency to G-PCC.

% Components	Geometry 	G-PCC Color	AV1 Color
% Decoding Time (sec)	0.656	2.15	0.254

\begin{table}
    \centering
    \caption{Decoding time (second) for one point cloud with different methods for geoemetry and texture, respectively. The decoding time is measured on a computer with Intel i7-9700K CPU and NVIDIA RTX 4080 GPU. G-PCC and AV1 use CPU for decoding. Results are averaged over the 4 scenes in the 8iVFB dataset.}
    \label{tab:decoding_time}
    
    \begin{tabular}{l|c|cc}
        \toprule
         & Geometry & \multicolumn{2}{c}{Texture} \\
        \midrule
        \multirow{2}{*}{Textured Surfel Octree} & \multirow{2}{*}{0.7} & G-PCC Color & AV1 Color\\
        & & 2.2 & 0.3 \\
        G-PCC (r03) & 1.1 & \multicolumn{2}{c}{1.0} \\
        G-PCC (r04) & 2.4 & \multicolumn{2}{c}{2.2} \\
        G-PCC (r05) & 3.1 & \multicolumn{2}{c}{2.9} \\
        \bottomrule
        \end{tabular}
\end{table}

\begin{figure*}
    \centering
    \small
    \begin{subfigure}{0.325\linewidth}
        \centering
        \includegraphics[width=\linewidth]{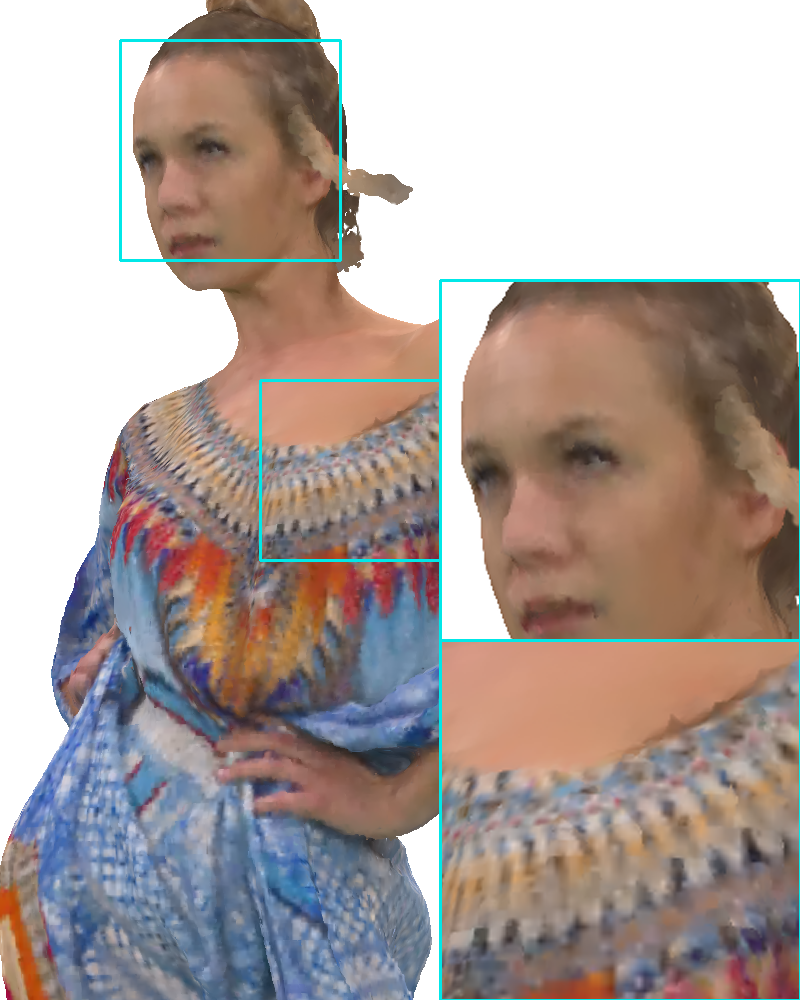}
    \end{subfigure}
    \begin{subfigure}{0.325\linewidth}
        \centering
        \includegraphics[width=\linewidth]{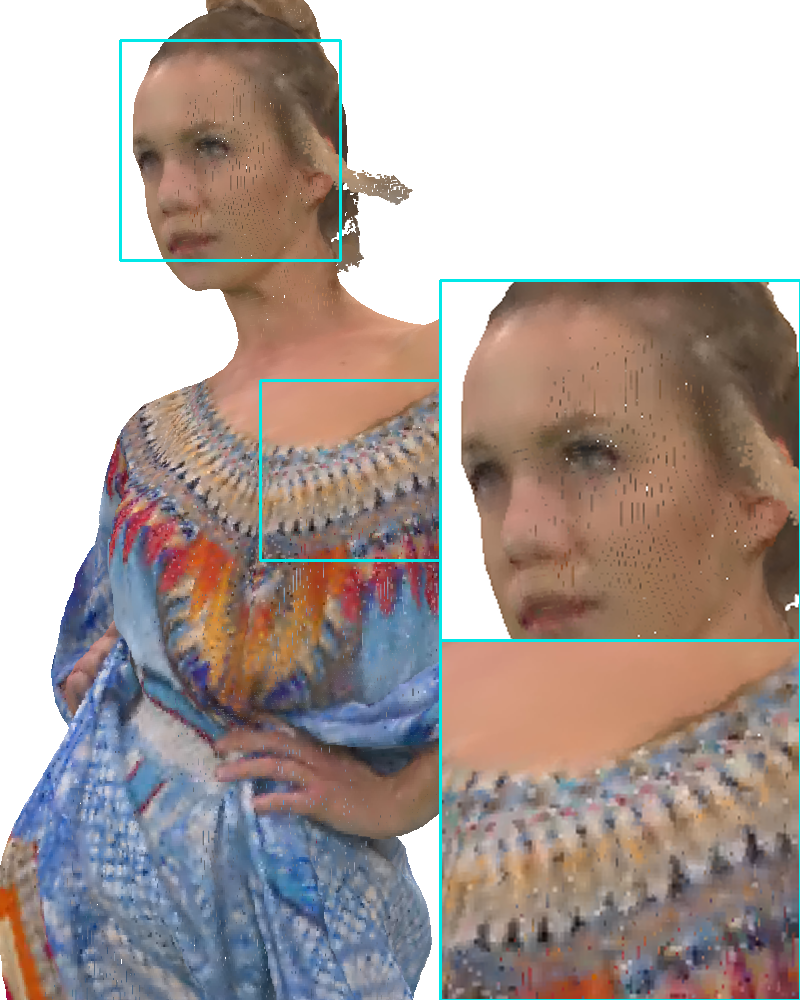}
    \end{subfigure}
    \begin{subfigure}{0.325\linewidth}
        \centering
        \includegraphics[width=\linewidth]{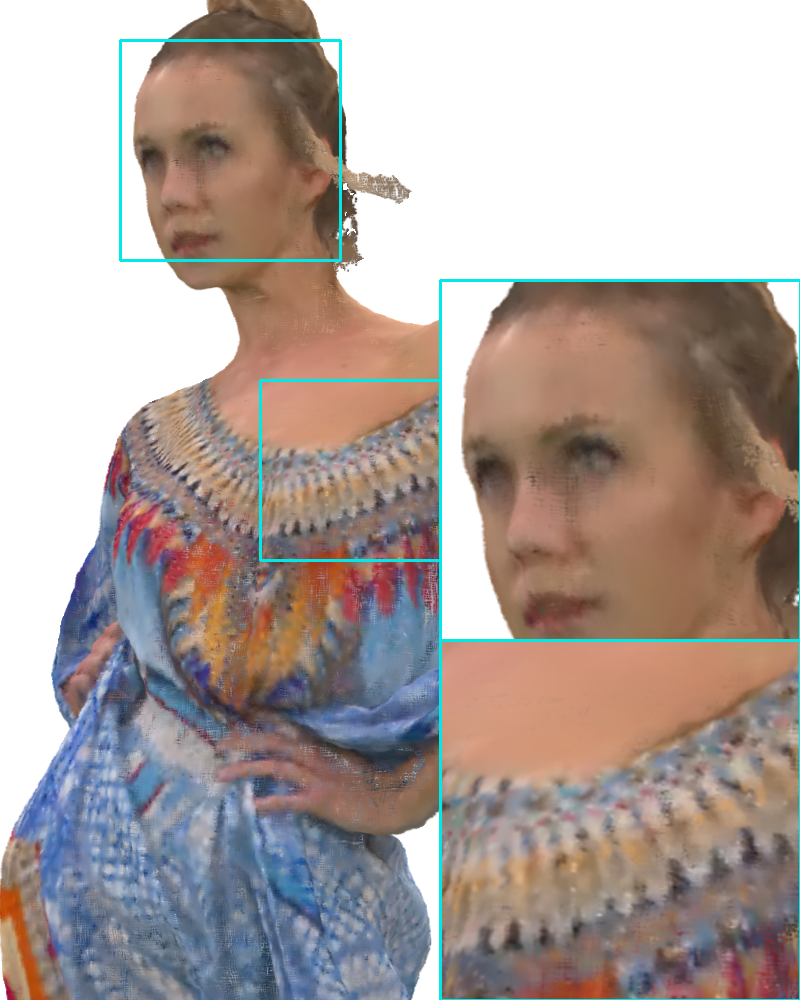}
    \end{subfigure}

    \begin{subfigure}{0.325\linewidth}
        \centering
        TeSO\\
        1.08 bpp
    \end{subfigure}
    \begin{subfigure}{0.325\linewidth}
        \centering
        G-PCC (r05) + OpenGL \\
        1.08 bpp
    \end{subfigure}
    \begin{subfigure}{0.325\linewidth}
        \centering
        G-PCC (r05) + P2ENet \\
        1.08 bpp
    \end{subfigure}

    \begin{subfigure}{0.325\linewidth}
      \centering
      \includegraphics[width=\linewidth]{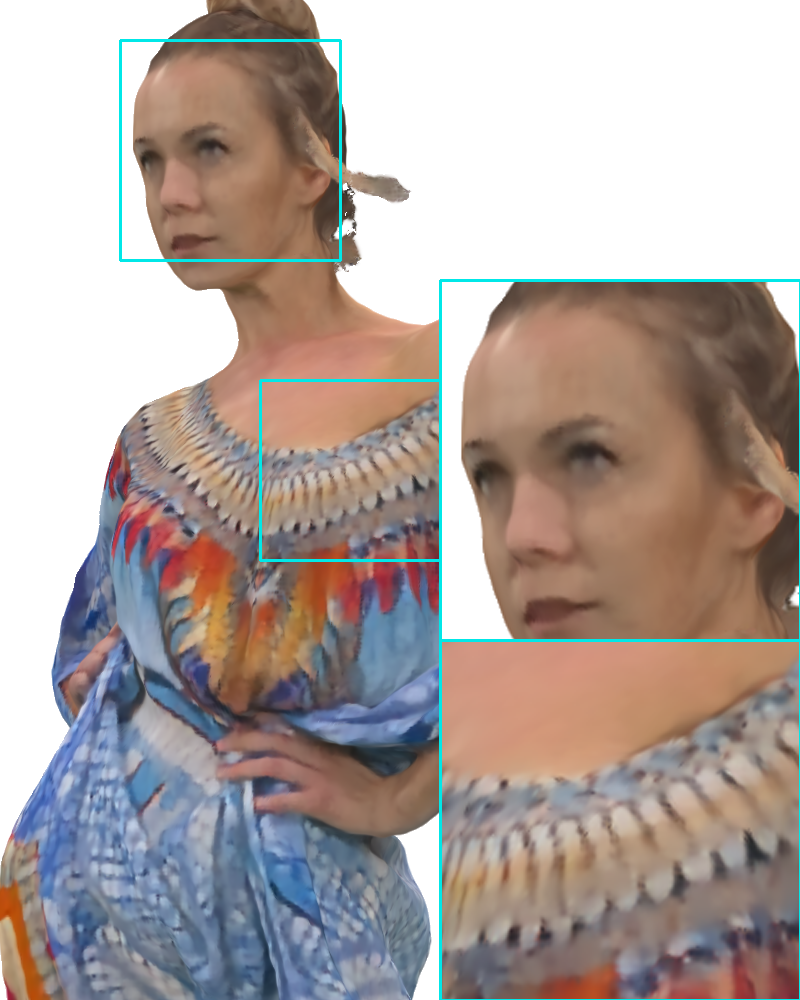}
  \end{subfigure}
  \begin{subfigure}{0.325\linewidth}
      \centering
      \includegraphics[width=\linewidth]{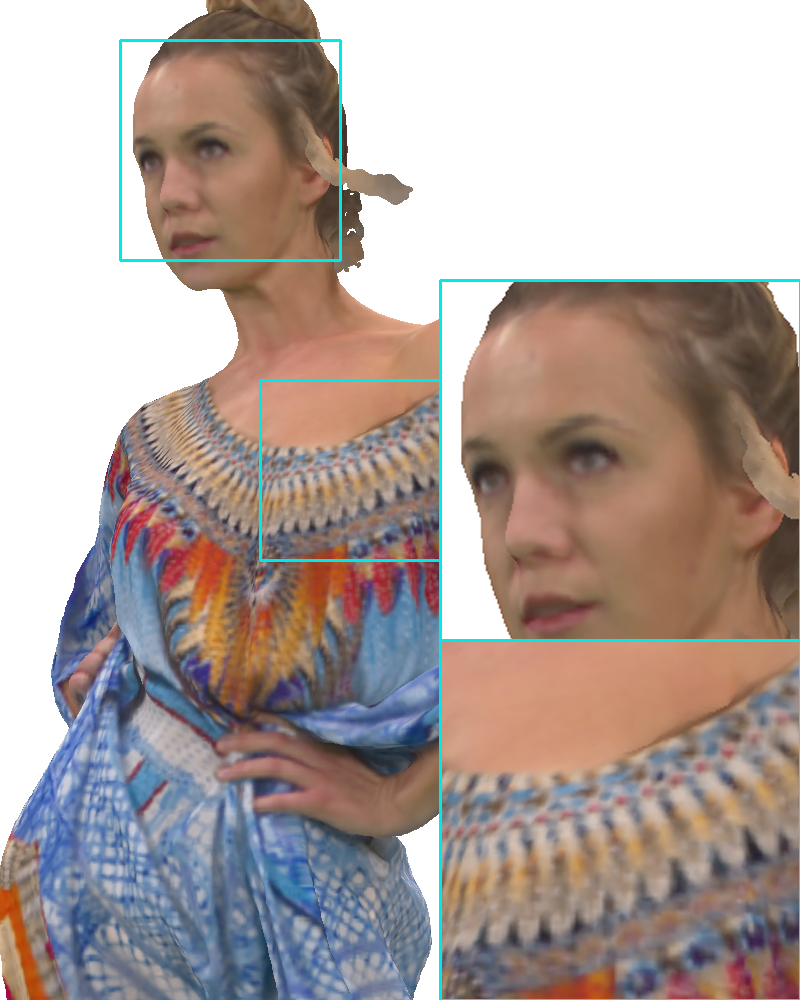}
  \end{subfigure}
  \vspace{2mm}

  \begin{subfigure}{0.325\linewidth}
    \centering
    B2P \\
    1.19 bpp
\end{subfigure}
\begin{subfigure}{0.325\linewidth}
    \centering
    Poisson Mesh (GT)
\end{subfigure}

    \caption{Rendering results of the compressed \textit{longdress} scene at resolution $1920 \times 1920$. The G-PCC decoded point clouds rendered by OpenGL has visible gaps. P2ENet achieves better quality but still has gap issues at high resolutions. B2P renders smooth results but suffers from  oversmoothing in the densely textured area. }\label{fig:longdress}
\end{figure*}

\begin{figure}
    \centering
    \small
    \begin{subfigure}{0.48\linewidth}
        \centering
        \includegraphics[width=1\linewidth]{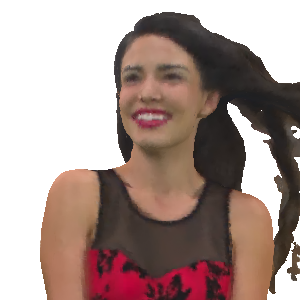}
    \end{subfigure}
    \begin{subfigure}{0.48\linewidth}
        \centering
        \includegraphics[width=1\linewidth]{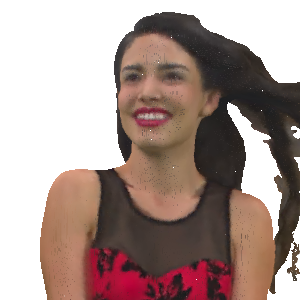}
    \end{subfigure}
    \vspace{2mm}

    \begin{subfigure}{0.48\linewidth}
        \centering
        TeSO\\
        1.25 bpp
    \end{subfigure}
    \begin{subfigure}{0.48\linewidth}
        \centering
        G-PCC (r05) + OpenGL \\
        1.29 bpp
    \end{subfigure}
    \caption{Rendering results of the \textit{redandblack}.}\label{fig:redandblack}
\end{figure}

\begin{figure}
    \centering
    \small
    \begin{subfigure}{0.48\linewidth}
        \centering
        \includegraphics[width=1\linewidth]{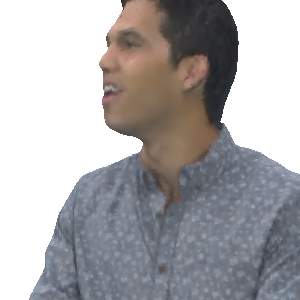}
    \end{subfigure}
    \begin{subfigure}{0.48\linewidth}
        \centering
        \includegraphics[width=1\linewidth]{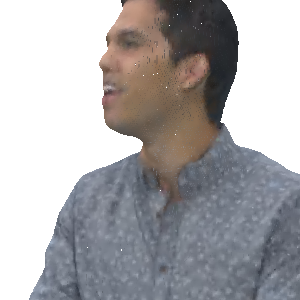}
    \end{subfigure}
    \vspace{2mm}

    \begin{subfigure}{0.48\linewidth}
        \centering
        TeSO\\
        0.55 bpp
    \end{subfigure}
    \begin{subfigure}{0.48\linewidth}
        \centering
        G-PCC (r04) + OpenGL \\
        0.55 bpp
    \end{subfigure}
    \caption{Rendering results of the \textit{Loot}. }\label{fig:loot}
\end{figure}

\begin{figure}
    \centering
    \small
    \begin{subfigure}{0.48\linewidth}
        \centering
        \includegraphics[width=1\linewidth]{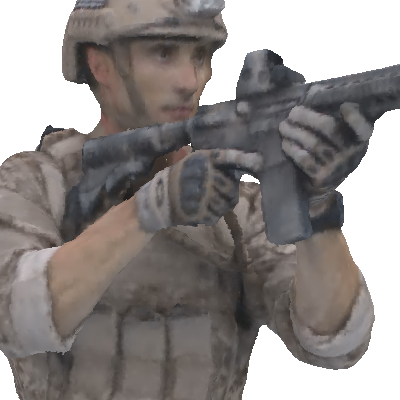}
    \end{subfigure}
    \begin{subfigure}{0.48\linewidth}
        \centering
        \includegraphics[width=1\linewidth]{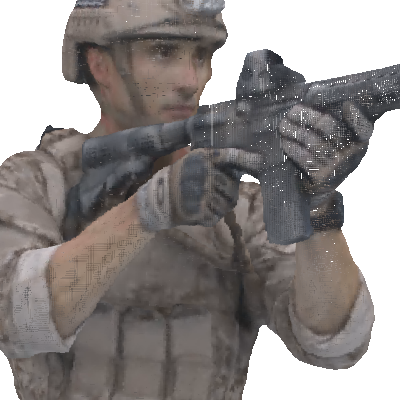}
    \end{subfigure}

    \begin{subfigure}{0.48\linewidth}
        \centering
        TeSO\\
        0.82 bpp
    \end{subfigure}
    \begin{subfigure}{0.48\linewidth}
        \centering
        G-PCC (r04) + OpenGL \\
        0.89 bpp
    \end{subfigure}
    \caption{Rendering results of the \textit{soldier}. }\label{fig:soldier}
\end{figure}

As shown in Fig.~\ref{fig:longdress}, the proposed compression and rendering pipeline can render smooth surface with dense high-frequency textures from decoded TeSOs. Due to the point-based rendering mechanism, OpenGL renders visible gaps, causing severe visual distortion. This is also illustrated in Fig.~\ref{fig:redandblack} to \ref{fig:soldier}. Although P2ENet improves the rendering quality, the rendered views still show point distortion at high resolutions. B2P renders smooth gap-free views but it still suffers from over-smoothing in dense texture areas. TeSO is more robust in rendering with dynamic cameras because it models explicit surface, making it more suitable for XR applications where the users' viewports are changing dynamically.

\section{Conclusion}
\label{sec:conclusion}

In this paper, we introduced TeSO, a novel 3D representation tailored for the demands of emerging XR applications and 3D visual content streaming. Recognizing the limitations of existing representations such as point clouds and meshes in balancing rendering quality, efficiency, and compression, we proposed a representation that combines the strengths of both. TeSO utilizes cube-bounded surfels organized within an octree, where each surfel is endowed with a texture patch. This design allows smooth surfaces to be represented compactly with large surfels, while preserving high-frequency visual details through the texture maps. We presented an efficient, GPU-accelerated algorithm to construct TeSO directly from point clouds. Furthermore, we developed a compression scheme specifically for this representation, employing a learned entropy model for the octree structure and geometry attributes, and leveraging standard video or point cloud color codecs for the texture patches.

We compare the proposed TeSO to the standard G-PCC point cloud compression combined with both traditional OpenGL and an advanced learned renderer (P2ENet), as well as the end-to-end optimized B2P method. Our approach achieves a superior trade-off between bit-rate and rendering quality, as measured by LPIPS, across various rendering resolutions. Visual comparisons confirmed the ability of our method to render sharp textures even with simplified geometry.

While TeSO shows significant promise, we also acknowledge limitations and areas for future work. Its current design is best suited for solid surfaces and may require extensions, possibly hybrid approaches combining surfels with other primitives like 3D Gaussians, to better handle complex phenomena such as hair or transparency. Additionally, in the experiments, we rely on LPIPS to measure the distortion assuming its alignment with human perception, to avoid the problem of over sensitivity of pixel-wise distortion metrics (\textit{i.e.} PSNR, SSIM). Nevertheless, a better metric that can measure the rendering quality of 3D representations is still an open problem and requires further research on human 3D perception. Finally, given the differentiability of the TeSO rendering, it is possible to design learning-based approach for  geometry and texture compression, to optimize for the rendering-quality vs. rate tradeoff.   Overall, TeSO offers a compelling new direction for representing and compressing 3D scenes, paving the way for more efficient and visually immersive streaming experiences.

%%
%% The next two lines define the bibliography style to be used, and
%% the bibliography file.
\bibliographystyle{ACM-Reference-Format}
\bibliography{main}

%%% -*-BibTeX-*-
%%% Do NOT edit. File created by BibTeX with style
%%% ACM-Reference-Format-Journals [18-Jan-2012].

\begin{thebibliography}{60}

%%% ====================================================================
%%% NOTE TO THE USER: you can override these defaults by providing
%%% customized versions of any of these macros before the \bibliography
%%% command.  Each of them MUST provide its own final punctuation,
%%% except for \shownote{} and \showURL{}.  The latter two
%%% do not use final punctuation, in order to avoid confusing it with
%%% the Web address.
%%%
%%% To suppress output of a particular field, define its macro to expand
%%% to an empty string, or better, \unskip, like this:
%%%
%%% \newcommand{\showURL}[1]{\unskip}   % LaTeX syntax
%%%
%%% \def \showURL #1{\unskip}           % plain TeX syntax
%%%
%%% ====================================================================

\ifx \showCODEN    \undefined \def \showCODEN     #1{\unskip}     \fi
\ifx \showISBNx    \undefined \def \showISBNx     #1{\unskip}     \fi
\ifx \showISBNxiii \undefined \def \showISBNxiii  #1{\unskip}     \fi
\ifx \showISSN     \undefined \def \showISSN      #1{\unskip}     \fi
\ifx \showLCCN     \undefined \def \showLCCN      #1{\unskip}     \fi
\ifx \shownote     \undefined \def \shownote      #1{#1}          \fi
\ifx \showarticletitle \undefined \def \showarticletitle #1{#1}   \fi
\ifx \showURL      \undefined \def \showURL       {\relax}        \fi
% The following commands are used for tagged output and should be
% invisible to TeX
\providecommand\bibfield[2]{#2}
\providecommand\bibinfo[2]{#2}
\providecommand\natexlab[1]{#1}
\providecommand\showeprint[2][]{arXiv:#2}

\bibitem[ope(2020)]%
        {openglprimitive}
 \bibinfo{year}{2020}\natexlab{}.
\newblock \bibinfo{booktitle}{\emph{OpenGL Primitive}}.
\newblock
\urldef\tempurl%
\url{https://www.khronos.org/opengl/wiki/primitive}
\showURL{%
\tempurl}
\newblock
\shownote{Accessed on April 22, 2025.}.


\bibitem[tmc(2023a)]%
        {tmc13}
 \bibinfo{year}{2023}\natexlab{a}.
\newblock \bibinfo{booktitle}{\emph{MPEG G-PCC TMC13}}.
\newblock
\urldef\tempurl%
\url{https://github.com/MPEGGroup/mpeg-pcc-tmc13}
\showURL{%
\tempurl}
\newblock
\shownote{Accessed on April 22, 2025.}.


\bibitem[tmc(2023b)]%
        {tmc2}
 \bibinfo{year}{2023}\natexlab{b}.
\newblock \bibinfo{booktitle}{\emph{MPEG V-PCC TMC2}}.
\newblock
\urldef\tempurl%
\url{https://github.com/MPEGGroup/mpeg-pcc-tmc2}
\showURL{%
\tempurl}
\newblock
\shownote{Accessed on April 22, 2025.}.


\bibitem[Aliev et~al\mbox{.}(2020)]%
        {aliev2020neural}
\bibfield{author}{\bibinfo{person}{Kara-Ali Aliev}, \bibinfo{person}{Artem Sevastopolsky}, \bibinfo{person}{Maria Kolos}, \bibinfo{person}{Dmitry Ulyanov}, {and} \bibinfo{person}{Victor Lempitsky}.} \bibinfo{year}{2020}\natexlab{}.
\newblock \showarticletitle{Neural point-based graphics}. In \bibinfo{booktitle}{\emph{ECCV}}. Springer, \bibinfo{pages}{696--712}.
\newblock


\bibitem[Bird et~al\mbox{.}(2021)]%
        {bird20213d}
\bibfield{author}{\bibinfo{person}{Thomas Bird}, \bibinfo{person}{Johannes Ball{\'e}}, \bibinfo{person}{Saurabh Singh}, {and} \bibinfo{person}{Philip~A Chou}.} \bibinfo{year}{2021}\natexlab{}.
\newblock \showarticletitle{3d scene compression through entropy penalized neural representation functions}. In \bibinfo{booktitle}{\emph{2021 Picture Coding Symposium (PCS)}}. IEEE, \bibinfo{pages}{1--5}.
\newblock


\bibitem[Chang et~al\mbox{.}(2023)]%
        {chang2023pointersect}
\bibfield{author}{\bibinfo{person}{Jen-Hao~Rick Chang}, \bibinfo{person}{Wei-Yu Chen}, \bibinfo{person}{Anurag Ranjan}, \bibinfo{person}{Kwang~Moo Yi}, {and} \bibinfo{person}{Oncel Tuzel}.} \bibinfo{year}{2023}\natexlab{}.
\newblock \showarticletitle{Pointersect: Neural Rendering with Cloud-Ray Intersection}. In \bibinfo{booktitle}{\emph{Proceedings of the IEEE/CVF Conference on Computer Vision and Pattern Recognition}}. \bibinfo{pages}{8359--8369}.
\newblock


\bibitem[Chen et~al\mbox{.}(2023)]%
        {chen2023introduction}
\bibfield{author}{\bibinfo{person}{Anthony Chen}, \bibinfo{person}{Shiwen Mao}, \bibinfo{person}{Zhu Li}, \bibinfo{person}{Minrui Xu}, \bibinfo{person}{Hongliang Zhang}, \bibinfo{person}{Dusit Niyato}, {and} \bibinfo{person}{Zhu Han}.} \bibinfo{year}{2023}\natexlab{}.
\newblock \showarticletitle{An Introduction to Point Cloud Compression Standards}.
\newblock \bibinfo{journal}{\emph{GetMobile: Mobile Computing and Communications}} \bibinfo{volume}{27}, \bibinfo{number}{1} (\bibinfo{year}{2023}), \bibinfo{pages}{11--17}.
\newblock


\bibitem[Chibane et~al\mbox{.}(2020)]%
        {chibane2020neural}
\bibfield{author}{\bibinfo{person}{Julian Chibane}, \bibinfo{person}{Gerard Pons-Moll}, {et~al\mbox{.}}} \bibinfo{year}{2020}\natexlab{}.
\newblock \showarticletitle{Neural unsigned distance fields for implicit function learning}.
\newblock \bibinfo{journal}{\emph{Advances in Neural Information Processing Systems}}  \bibinfo{volume}{33} (\bibinfo{year}{2020}), \bibinfo{pages}{21638--21652}.
\newblock


\bibitem[Cui et~al\mbox{.}(2023)]%
        {cui2023octformer}
\bibfield{author}{\bibinfo{person}{Mingyue Cui}, \bibinfo{person}{Junhua Long}, \bibinfo{person}{Mingjian Feng}, \bibinfo{person}{Boyang Li}, {and} \bibinfo{person}{Huang Kai}.} \bibinfo{year}{2023}\natexlab{}.
\newblock \showarticletitle{OctFormer: Efficient octree-based transformer for point cloud compression with local enhancement}. In \bibinfo{booktitle}{\emph{Proceedings of the AAAI Conference on Artificial Intelligence}}, Vol.~\bibinfo{volume}{37}. \bibinfo{pages}{470--478}.
\newblock


\bibitem[d'Eon et~al\mbox{.}(2017)]%
        {dataset8i}
\bibfield{author}{\bibinfo{person}{Eugene d'Eon}, \bibinfo{person}{Harrison Bob}, \bibinfo{person}{Taos Myers}, {and} \bibinfo{person}{Philip~A. Chou}.} \bibinfo{year}{2017}\natexlab{}.
\newblock \showarticletitle{8i Voxelized Full Bodies - A Voxelized Point Cloud Dataset}. In \bibinfo{booktitle}{\emph{ISO/IEC JTC1/SC29 Joint WG11/WG1 (MPEG/JPEG) input document WG11M40059/WG1M74006}}.
\newblock


\bibitem[Fu et~al\mbox{.}(2022)]%
        {fu2022octattention}
\bibfield{author}{\bibinfo{person}{Chunyang Fu}, \bibinfo{person}{Ge Li}, \bibinfo{person}{Rui Song}, \bibinfo{person}{Wei Gao}, {and} \bibinfo{person}{Shan Liu}.} \bibinfo{year}{2022}\natexlab{}.
\newblock \showarticletitle{Octattention: Octree-based large-scale contexts model for point cloud compression}. In \bibinfo{booktitle}{\emph{Proceedings of the AAAI Conference on Artificial Intelligence}}, Vol.~\bibinfo{volume}{36}. \bibinfo{pages}{625--633}.
\newblock


\bibitem[Funk et~al\mbox{.}(2017)]%
        {funk2017holocollab}
\bibfield{author}{\bibinfo{person}{Markus Funk}, \bibinfo{person}{Mareike Kritzler}, {and} \bibinfo{person}{Florian Michahelles}.} \bibinfo{year}{2017}\natexlab{}.
\newblock \showarticletitle{HoloCollab: a shared virtual platform for physical assembly training using spatially-aware head-mounted displays}. In \bibinfo{booktitle}{\emph{Proceedings of the Seventh International Conference on the Internet of Things}}. \bibinfo{pages}{1--7}.
\newblock


\bibitem[Gautier et~al\mbox{.}(2023)]%
        {gautier2023uvg}
\bibfield{author}{\bibinfo{person}{Guillaume Gautier}, \bibinfo{person}{Alexandre Mercat}, \bibinfo{person}{Louis Fr{\'e}neau}, \bibinfo{person}{Mikko Pitk{\"a}nen}, {and} \bibinfo{person}{Jarno Vanne}.} \bibinfo{year}{2023}\natexlab{}.
\newblock \showarticletitle{UVG-VPC: voxelized point cloud dataset for visual volumetric video-based coding}. In \bibinfo{booktitle}{\emph{2023 15th international conference on quality of Multimedia experience (QoMEX)}}. IEEE, \bibinfo{pages}{244--247}.
\newblock


\bibitem[Graziosi et~al\mbox{.}(2020)]%
        {graziosi2020overview}
\bibfield{author}{\bibinfo{person}{Danillo Graziosi}, \bibinfo{person}{Ohji Nakagami}, \bibinfo{person}{Shinroku Kuma}, \bibinfo{person}{Alexandre Zaghetto}, \bibinfo{person}{Teruhiko Suzuki}, {and} \bibinfo{person}{Ali Tabatabai}.} \bibinfo{year}{2020}\natexlab{}.
\newblock \showarticletitle{An overview of ongoing point cloud compression standardization activities: Video-based (V-PCC) and geometry-based (G-PCC)}.
\newblock \bibinfo{journal}{\emph{APSIPA Transactions on Signal and Information Processing}}  \bibinfo{volume}{9} (\bibinfo{year}{2020}), \bibinfo{pages}{e13}.
\newblock


\bibitem[Guan et~al\mbox{.}(2023)]%
        {guan2023metastream}
\bibfield{author}{\bibinfo{person}{Yongjie Guan}, \bibinfo{person}{Xueyu Hou}, \bibinfo{person}{Nan Wu}, \bibinfo{person}{Bo Han}, {and} \bibinfo{person}{Tao Han}.} \bibinfo{year}{2023}\natexlab{}.
\newblock \showarticletitle{Metastream: Live volumetric content capture, creation, delivery, and rendering in real time}. In \bibinfo{booktitle}{\emph{Proceedings of the 29th Annual International Conference on Mobile Computing and Networking}}. \bibinfo{pages}{1--15}.
\newblock


\bibitem[Han et~al\mbox{.}(2021)]%
        {han2021technical}
\bibfield{author}{\bibinfo{person}{Jingning Han}, \bibinfo{person}{Bohan Li}, \bibinfo{person}{Debargha Mukherjee}, \bibinfo{person}{Ching-Han Chiang}, \bibinfo{person}{Adrian Grange}, \bibinfo{person}{Cheng Chen}, \bibinfo{person}{Hui Su}, \bibinfo{person}{Sarah Parker}, \bibinfo{person}{Sai Deng}, \bibinfo{person}{Urvang Joshi}, {et~al\mbox{.}}} \bibinfo{year}{2021}\natexlab{}.
\newblock \showarticletitle{A technical overview of AV1}.
\newblock \bibinfo{journal}{\emph{Proc. IEEE}} \bibinfo{volume}{109}, \bibinfo{number}{9} (\bibinfo{year}{2021}), \bibinfo{pages}{1435--1462}.
\newblock


\bibitem[He et~al\mbox{.}(2022)]%
        {he2022density}
\bibfield{author}{\bibinfo{person}{Yun He}, \bibinfo{person}{Xinlin Ren}, \bibinfo{person}{Danhang Tang}, \bibinfo{person}{Yinda Zhang}, \bibinfo{person}{Xiangyang Xue}, {and} \bibinfo{person}{Yanwei Fu}.} \bibinfo{year}{2022}\natexlab{}.
\newblock \showarticletitle{Density-preserving deep point cloud compression}. In \bibinfo{booktitle}{\emph{CVPR}}.
\newblock


\bibitem[Hoppe et~al\mbox{.}(1992)]%
        {hoppe1992surface}
\bibfield{author}{\bibinfo{person}{Hugues Hoppe}, \bibinfo{person}{Tony DeRose}, \bibinfo{person}{Tom Duchamp}, \bibinfo{person}{John McDonald}, {and} \bibinfo{person}{Werner Stuetzle}.} \bibinfo{year}{1992}\natexlab{}.
\newblock \showarticletitle{Surface reconstruction from unorganized points}. In \bibinfo{booktitle}{\emph{Proceedings of the 19th annual conference on computer graphics and interactive techniques}}. \bibinfo{pages}{71--78}.
\newblock


\bibitem[Hu et~al\mbox{.}(2024b)]%
        {hu2024low}
\bibfield{author}{\bibinfo{person}{Yueyu Hu}, \bibinfo{person}{Ran Gong}, \bibinfo{person}{Qi Sun}, {and} \bibinfo{person}{Yao Wang}.} \bibinfo{year}{2024}\natexlab{b}.
\newblock \showarticletitle{Low Latency Point Cloud Rendering with Learned Splatting}. In \bibinfo{booktitle}{\emph{Proceedings of the IEEE/CVF Conference on Computer Vision and Pattern Recognition}}. \bibinfo{pages}{5752--5761}.
\newblock


\bibitem[Hu et~al\mbox{.}(2024a)]%
        {hu2024bits}
\bibfield{author}{\bibinfo{person}{Yueyu Hu}, \bibinfo{person}{Ran Gong}, {and} \bibinfo{person}{Yao Wang}.} \bibinfo{year}{2024}\natexlab{a}.
\newblock \showarticletitle{Bits-to-photon: End-to-end learned scalable point cloud compression for direct rendering}.
\newblock \bibinfo{journal}{\emph{arXiv preprint arXiv:2406.05915}} (\bibinfo{year}{2024}).
\newblock


\bibitem[Huang et~al\mbox{.}(2020)]%
        {huang2020octsqueeze}
\bibfield{author}{\bibinfo{person}{Lila Huang}, \bibinfo{person}{Shenlong Wang}, \bibinfo{person}{Kelvin Wong}, \bibinfo{person}{Jerry Liu}, {and} \bibinfo{person}{Raquel Urtasun}.} \bibinfo{year}{2020}\natexlab{}.
\newblock \showarticletitle{Octsqueeze: Octree-structured entropy model for lidar compression}. In \bibinfo{booktitle}{\emph{CVPR}}. \bibinfo{pages}{1313--1323}.
\newblock


\bibitem[Jin et~al\mbox{.}(2024)]%
        {jin20243d}
\bibfield{author}{\bibinfo{person}{Yili Jin}, \bibinfo{person}{Xize Duan}, \bibinfo{person}{Kaiyuan Hu}, \bibinfo{person}{Fangxin Wang}, {and} \bibinfo{person}{Xue Liu}.} \bibinfo{year}{2024}\natexlab{}.
\newblock \showarticletitle{3D Video Conferencing via On-hand Devices}.
\newblock \bibinfo{journal}{\emph{IEEE Transactions on Circuits and Systems for Video Technology}} (\bibinfo{year}{2024}).
\newblock


\bibitem[Kami{\'n}ska et~al\mbox{.}(2023)]%
        {kaminska2023augmented}
\bibfield{author}{\bibinfo{person}{Dorota Kami{\'n}ska}, \bibinfo{person}{Grzegorz Zwoli{\'n}ski}, \bibinfo{person}{Anna Laska-Le{\'s}niewicz}, \bibinfo{person}{Rui Raposo}, \bibinfo{person}{M{\'a}rio Vairinhos}, \bibinfo{person}{Elisabeth Pereira}, \bibinfo{person}{Frane Urem}, \bibinfo{person}{Martina Ljubi{\'c}~Hini{\'c}}, \bibinfo{person}{Rain~Eric Haamer}, {and} \bibinfo{person}{Gholamreza Anbarjafari}.} \bibinfo{year}{2023}\natexlab{}.
\newblock \showarticletitle{Augmented reality: Current and new trends in education}.
\newblock \bibinfo{journal}{\emph{Electronics}} \bibinfo{volume}{12}, \bibinfo{number}{16} (\bibinfo{year}{2023}), \bibinfo{pages}{3531}.
\newblock


\bibitem[Kazhdan and Hoppe(2013)]%
        {kazhdan2013screened}
\bibfield{author}{\bibinfo{person}{Michael Kazhdan} {and} \bibinfo{person}{Hugues Hoppe}.} \bibinfo{year}{2013}\natexlab{}.
\newblock \showarticletitle{Screened poisson surface reconstruction}.
\newblock \bibinfo{journal}{\emph{TOG}} \bibinfo{volume}{32}, \bibinfo{number}{3} (\bibinfo{year}{2013}), \bibinfo{pages}{1--13}.
\newblock


\bibitem[Kerbl et~al\mbox{.}(2023)]%
        {kerbl20233d}
\bibfield{author}{\bibinfo{person}{Bernhard Kerbl}, \bibinfo{person}{Georgios Kopanas}, \bibinfo{person}{Thomas Leimk{\"u}hler}, {and} \bibinfo{person}{George Drettakis}.} \bibinfo{year}{2023}\natexlab{}.
\newblock \showarticletitle{3D Gaussian Splatting for Real-Time Radiance Field Rendering}.
\newblock \bibinfo{journal}{\emph{ACM Transactions on Graphics}} \bibinfo{volume}{42}, \bibinfo{number}{4} (\bibinfo{year}{2023}).
\newblock


\bibitem[Kobbelt and Botsch(2004)]%
        {kobbelt2004survey}
\bibfield{author}{\bibinfo{person}{Leif Kobbelt} {and} \bibinfo{person}{Mario Botsch}.} \bibinfo{year}{2004}\natexlab{}.
\newblock \showarticletitle{A survey of point-based techniques in computer graphics}.
\newblock \bibinfo{journal}{\emph{Computers \& Graphics}} \bibinfo{volume}{28}, \bibinfo{number}{6} (\bibinfo{year}{2004}), \bibinfo{pages}{801--814}.
\newblock


\bibitem[Lawrence et~al\mbox{.}(2024)]%
        {lawrence2024project}
\bibfield{author}{\bibinfo{person}{Jason Lawrence}, \bibinfo{person}{Ryan Overbeck}, \bibinfo{person}{Todd Prives}, \bibinfo{person}{Tommy Fortes}, \bibinfo{person}{Nikki Roth}, {and} \bibinfo{person}{Brett Newman}.} \bibinfo{year}{2024}\natexlab{}.
\newblock \showarticletitle{Project starline: A high-fidelity telepresence system}.
\newblock In \bibinfo{booktitle}{\emph{ACM SIGGRAPH 2024 Emerging Technologies}}. \bibinfo{pages}{1--2}.
\newblock


\bibitem[Lee et~al\mbox{.}(2020)]%
        {lee2020groot}
\bibfield{author}{\bibinfo{person}{Kyungjin Lee}, \bibinfo{person}{Juheon Yi}, \bibinfo{person}{Youngki Lee}, \bibinfo{person}{Sunghyun Choi}, {and} \bibinfo{person}{Young~Min Kim}.} \bibinfo{year}{2020}\natexlab{}.
\newblock \showarticletitle{GROOT: a real-time streaming system of high-fidelity volumetric videos}. In \bibinfo{booktitle}{\emph{Proceedings of the 26th Annual International Conference on Mobile Computing and Networking}}. \bibinfo{pages}{1--14}.
\newblock


\bibitem[Li et~al\mbox{.}(2024b)]%
        {li2024spatial}
\bibfield{author}{\bibinfo{person}{Chen Li}, \bibinfo{person}{Tongyu Zong}, \bibinfo{person}{Yueyu Hu}, \bibinfo{person}{Yao Wang}, {and} \bibinfo{person}{Yong Liu}.} \bibinfo{year}{2024}\natexlab{b}.
\newblock \showarticletitle{Spatial Visibility and Temporal Dynamics: Revolutionizing Field of View Prediction in Adaptive Point Cloud Video Streaming}.
\newblock \bibinfo{journal}{\emph{arXiv preprint arXiv:2409.18236}} (\bibinfo{year}{2024}).
\newblock


\bibitem[Li et~al\mbox{.}(2024a)]%
        {li2024learning}
\bibfield{author}{\bibinfo{person}{Qing Li}, \bibinfo{person}{Huifang Feng}, \bibinfo{person}{Kanle Shi}, \bibinfo{person}{Yue Gao}, \bibinfo{person}{Yi Fang}, \bibinfo{person}{Yu-Shen Liu}, {and} \bibinfo{person}{Zhizhong Han}.} \bibinfo{year}{2024}\natexlab{a}.
\newblock \showarticletitle{Learning signed hyper surfaces for oriented point cloud normal estimation}.
\newblock \bibinfo{journal}{\emph{IEEE Transactions on Pattern Analysis and Machine Intelligence}} (\bibinfo{year}{2024}).
\newblock


\bibitem[Liang et~al\mbox{.}(2024)]%
        {liang2024fumos}
\bibfield{author}{\bibinfo{person}{Zhicheng Liang}, \bibinfo{person}{Junhua Liu}, \bibinfo{person}{Mallesham Dasari}, {and} \bibinfo{person}{Fangxin Wang}.} \bibinfo{year}{2024}\natexlab{}.
\newblock \showarticletitle{Fumos: Neural compression and progressive refinement for continuous point cloud video streaming}.
\newblock \bibinfo{journal}{\emph{IEEE Transactions on Visualization and Computer Graphics}} (\bibinfo{year}{2024}).
\newblock


\bibitem[Lindell et~al\mbox{.}(2022)]%
        {lindell2022bacon}
\bibfield{author}{\bibinfo{person}{David~B Lindell}, \bibinfo{person}{Dave Van~Veen}, \bibinfo{person}{Jeong~Joon Park}, {and} \bibinfo{person}{Gordon Wetzstein}.} \bibinfo{year}{2022}\natexlab{}.
\newblock \showarticletitle{Bacon: Band-limited coordinate networks for multiscale scene representation}. In \bibinfo{booktitle}{\emph{Proceedings of the IEEE/CVF conference on computer vision and pattern recognition}}. \bibinfo{pages}{16252--16262}.
\newblock


\bibitem[Lionar et~al\mbox{.}(2023)]%
        {lionar2023nu}
\bibfield{author}{\bibinfo{person}{Stefan Lionar}, \bibinfo{person}{Xiangyu Xu}, \bibinfo{person}{Min Lin}, {and} \bibinfo{person}{Gim~Hee Lee}.} \bibinfo{year}{2023}\natexlab{}.
\newblock \showarticletitle{Nu-mcc: Multiview compressive coding with neighborhood decoder and repulsive udf}.
\newblock \bibinfo{journal}{\emph{Advances in Neural Information Processing Systems}}  \bibinfo{volume}{36} (\bibinfo{year}{2023}), \bibinfo{pages}{63011--63022}.
\newblock


\bibitem[Liu et~al\mbox{.}(2023)]%
        {liu2023cav3}
\bibfield{author}{\bibinfo{person}{Junhua Liu}, \bibinfo{person}{Boxiang Zhu}, \bibinfo{person}{Fangxin Wang}, \bibinfo{person}{Yili Jin}, \bibinfo{person}{Wenyi Zhang}, \bibinfo{person}{Zihan Xu}, {and} \bibinfo{person}{Shuguang Cui}.} \bibinfo{year}{2023}\natexlab{}.
\newblock \showarticletitle{Cav3: Cache-assisted viewport adaptive volumetric video streaming}. In \bibinfo{booktitle}{\emph{2023 IEEE Conference Virtual Reality and 3D User Interfaces (VR)}}. IEEE, \bibinfo{pages}{173--183}.
\newblock


\bibitem[Mao et~al\mbox{.}(2022)]%
        {mao2022learning}
\bibfield{author}{\bibinfo{person}{Yixiang Mao}, \bibinfo{person}{Yueyu Hu}, {and} \bibinfo{person}{Yao Wang}.} \bibinfo{year}{2022}\natexlab{}.
\newblock \showarticletitle{Learning to predict on octree for scalable point cloud geometry coding}. In \bibinfo{booktitle}{\emph{2022 IEEE 5th International Conference on Multimedia Information Processing and Retrieval (MIPR)}}. IEEE, \bibinfo{pages}{96--102}.
\newblock


\bibitem[Meyer et~al\mbox{.}(2010)]%
        {meyer2010floating}
\bibfield{author}{\bibinfo{person}{Quirin Meyer}, \bibinfo{person}{Jochen S{\"u}{\ss}muth}, \bibinfo{person}{Gerd Su{\ss}ner}, \bibinfo{person}{Marc Stamminger}, {and} \bibinfo{person}{G{\"u}nther Greiner}.} \bibinfo{year}{2010}\natexlab{}.
\newblock \showarticletitle{On floating-point normal vectors}. In \bibinfo{booktitle}{\emph{Computer Graphics Forum}}, Vol.~\bibinfo{volume}{29}. Wiley Online Library, \bibinfo{pages}{1405--1409}.
\newblock


\bibitem[Mildenhall et~al\mbox{.}(2021)]%
        {mildenhall2021nerf}
\bibfield{author}{\bibinfo{person}{Ben Mildenhall}, \bibinfo{person}{Pratul~P Srinivasan}, \bibinfo{person}{Matthew Tancik}, \bibinfo{person}{Jonathan~T Barron}, \bibinfo{person}{Ravi Ramamoorthi}, {and} \bibinfo{person}{Ren Ng}.} \bibinfo{year}{2021}\natexlab{}.
\newblock \showarticletitle{Nerf: Representing scenes as neural radiance fields for view synthesis}.
\newblock \bibinfo{journal}{\emph{Commun. ACM}} \bibinfo{volume}{65}, \bibinfo{number}{1} (\bibinfo{year}{2021}), \bibinfo{pages}{99--106}.
\newblock


\bibitem[Mühlenbrock et~al\mbox{.}(2024)]%
        {hasegawa2024blendpcr}
\bibfield{author}{\bibinfo{person}{Andre Mühlenbrock}, \bibinfo{person}{Rene Weller}, {and} \bibinfo{person}{Gabriel Zachmann}.} \bibinfo{year}{2024}\natexlab{}.
\newblock \showarticletitle{{BlendPCR: Seamless and Efficient Rendering of Dynamic Point Clouds captured by Multiple RGB-D Cameras}}. In \bibinfo{booktitle}{\emph{ICAT-EGVE 2024 - International Conference on Artificial Reality and Telexistence and Eurographics Symposium on Virtual Environments}}, \bibfield{editor}{\bibinfo{person}{Shoichi Hasegawa}, \bibinfo{person}{Nobuchika Sakata}, {and} \bibinfo{person}{Veronica Sundstedt}} (Eds.). \bibinfo{publisher}{The Eurographics Association}.
\newblock
\showISBNx{978-3-03868-245-5}
\showISSN{1727-530X}
\href{https://doi.org/10.2312/egve.20241366}{doi:\nolinkurl{10.2312/egve.20241366}}


\bibitem[Nakagami et~al\mbox{.}(2023)]%
        {gpccwhitepaper}
\bibfield{author}{\bibinfo{person}{Ohji Nakagami}, \bibinfo{person}{Sebastien Lasserre}, \bibinfo{person}{Sugio Toshiyasu}, {and} \bibinfo{person}{Marius Preda}.} \bibinfo{year}{2023}\natexlab{}.
\newblock \showarticletitle{White paper on G-PCC}. In \bibinfo{booktitle}{\emph{ISO/IEC JTC 1/SC 29/AG 03 N0111}}.
\newblock
\urldef\tempurl%
\url{https://www.mpeg.org/wp-content/uploads/mpeg_meetings/142_Antalya/w22804.zip}
\showURL{%
\tempurl}


\bibitem[Niedermayr et~al\mbox{.}(2024)]%
        {niedermayr2024compressed}
\bibfield{author}{\bibinfo{person}{Simon Niedermayr}, \bibinfo{person}{Josef Stumpfegger}, {and} \bibinfo{person}{R{\"u}diger Westermann}.} \bibinfo{year}{2024}\natexlab{}.
\newblock \showarticletitle{Compressed 3d gaussian splatting for accelerated novel view synthesis}. In \bibinfo{booktitle}{\emph{Proceedings of the IEEE/CVF Conference on Computer Vision and Pattern Recognition}}. \bibinfo{pages}{10349--10358}.
\newblock


\bibitem[Pfister et~al\mbox{.}(2000)]%
        {pfister2000surfels}
\bibfield{author}{\bibinfo{person}{Hanspeter Pfister}, \bibinfo{person}{Matthias Zwicker}, \bibinfo{person}{Jeroen Van~Baar}, {and} \bibinfo{person}{Markus Gross}.} \bibinfo{year}{2000}\natexlab{}.
\newblock \showarticletitle{Surfels: Surface elements as rendering primitives}. In \bibinfo{booktitle}{\emph{Proceedings of the 27th annual conference on Computer graphics and interactive techniques}}. \bibinfo{pages}{335--342}.
\newblock


\bibitem[Que et~al\mbox{.}(2021)]%
        {que2021voxelcontext}
\bibfield{author}{\bibinfo{person}{Zizheng Que}, \bibinfo{person}{Guo Lu}, {and} \bibinfo{person}{Dong Xu}.} \bibinfo{year}{2021}\natexlab{}.
\newblock \showarticletitle{Voxelcontext-net: An octree based framework for point cloud compression}. In \bibinfo{booktitle}{\emph{CVPR}}. \bibinfo{pages}{6042--6051}.
\newblock


\bibitem[Reipschl{\"a}ger and Dachselt(2019)]%
        {reipschlager2019designar}
\bibfield{author}{\bibinfo{person}{Patrick Reipschl{\"a}ger} {and} \bibinfo{person}{Raimund Dachselt}.} \bibinfo{year}{2019}\natexlab{}.
\newblock \showarticletitle{Designar: Immersive 3d-modeling combining augmented reality with interactive displays}. In \bibinfo{booktitle}{\emph{Proceedings of the 2019 ACM international conference on interactive surfaces and spaces}}. \bibinfo{pages}{29--41}.
\newblock


\bibitem[Schwarz et~al\mbox{.}(2021)]%
        {schwarz2021cabac}
\bibfield{author}{\bibinfo{person}{Heiko Schwarz}, \bibinfo{person}{Muhammed Coban}, \bibinfo{person}{Marta Karczewicz}, \bibinfo{person}{Tzu-Der Chuang}, \bibinfo{person}{Frank Bossen}, \bibinfo{person}{Alexander Alshin}, \bibinfo{person}{Jani Lainema}, \bibinfo{person}{Christian~R. Helmrich}, {and} \bibinfo{person}{Thomas Wiegand}.} \bibinfo{year}{2021}\natexlab{}.
\newblock \showarticletitle{Quantization and Entropy Coding in the Versatile Video Coding (VVC) Standard}.
\newblock \bibinfo{journal}{\emph{IEEE Transactions on Circuits and Systems for Video Technology}} \bibinfo{volume}{31}, \bibinfo{number}{10} (\bibinfo{year}{2021}), \bibinfo{pages}{3891--3906}.
\newblock


\bibitem[Stengel et~al\mbox{.}(2023)]%
        {stengel2023ai}
\bibfield{author}{\bibinfo{person}{Michael Stengel}, \bibinfo{person}{Koki Nagano}, \bibinfo{person}{Chao Liu}, \bibinfo{person}{Matthew Chan}, \bibinfo{person}{Alex Trevithick}, \bibinfo{person}{Shalini De~Mello}, \bibinfo{person}{Jonghyun Kim}, {and} \bibinfo{person}{David Luebke}.} \bibinfo{year}{2023}\natexlab{}.
\newblock \showarticletitle{Ai-mediated 3d video conferencing}.
\newblock In \bibinfo{booktitle}{\emph{ACM SIGGRAPH 2023 Emerging Technologies}}. \bibinfo{pages}{1--2}.
\newblock


\bibitem[Takikawa et~al\mbox{.}(2022)]%
        {takikawa2022variable}
\bibfield{author}{\bibinfo{person}{Towaki Takikawa}, \bibinfo{person}{Alex Evans}, \bibinfo{person}{Jonathan Tremblay}, \bibinfo{person}{Thomas M{\"u}ller}, \bibinfo{person}{Morgan McGuire}, \bibinfo{person}{Alec Jacobson}, {and} \bibinfo{person}{Sanja Fidler}.} \bibinfo{year}{2022}\natexlab{}.
\newblock \showarticletitle{Variable bitrate neural fields}. In \bibinfo{booktitle}{\emph{ACM SIGGRAPH 2022 Conference Proceedings}}. \bibinfo{pages}{1--9}.
\newblock


\bibitem[Tewari et~al\mbox{.}(2022)]%
        {tewari2022advances}
\bibfield{author}{\bibinfo{person}{Ayush Tewari}, \bibinfo{person}{Justus Thies}, \bibinfo{person}{Ben Mildenhall}, \bibinfo{person}{Pratul Srinivasan}, \bibinfo{person}{Edgar Tretschk}, \bibinfo{person}{Wang Yifan}, \bibinfo{person}{Christoph Lassner}, \bibinfo{person}{Vincent Sitzmann}, \bibinfo{person}{Ricardo Martin-Brualla}, \bibinfo{person}{Stephen Lombardi}, {et~al\mbox{.}}} \bibinfo{year}{2022}\natexlab{}.
\newblock \showarticletitle{Advances in neural rendering}. In \bibinfo{booktitle}{\emph{Computer Graphics Forum}}, Vol.~\bibinfo{volume}{41}. Wiley Online Library, \bibinfo{pages}{703--735}.
\newblock


\bibitem[Wang et~al\mbox{.}(2022)]%
        {wang2022sparsepcgc}
\bibfield{author}{\bibinfo{person}{Jianqiang Wang}, \bibinfo{person}{Dandan Ding}, \bibinfo{person}{Zhu Li}, \bibinfo{person}{Xiaoxing Feng}, \bibinfo{person}{Chuntong Cao}, {and} \bibinfo{person}{Zhan Ma}.} \bibinfo{year}{2022}\natexlab{}.
\newblock \showarticletitle{Sparse tensor-based multiscale representation for point cloud geometry compression}.
\newblock \bibinfo{journal}{\emph{IEEE Transactions on Pattern Analysis and Machine Intelligence}} (\bibinfo{year}{2022}).
\newblock


\bibitem[Wang et~al\mbox{.}(2024a)]%
        {wang2024versatile}
\bibfield{author}{\bibinfo{person}{Jianqiang Wang}, \bibinfo{person}{Ruixiang Xue}, \bibinfo{person}{Jiaxin Li}, \bibinfo{person}{Dandan Ding}, \bibinfo{person}{Yi Lin}, {and} \bibinfo{person}{Zhan Ma}.} \bibinfo{year}{2024}\natexlab{a}.
\newblock \showarticletitle{A versatile point cloud compressor using universal multiscale conditional coding--Part I: Geometry}.
\newblock \bibinfo{journal}{\emph{IEEE transactions on pattern analysis and machine intelligence}} (\bibinfo{year}{2024}).
\newblock


\bibitem[Wang et~al\mbox{.}(2024b)]%
        {wang2024versatile2}
\bibfield{author}{\bibinfo{person}{Jianqiang Wang}, \bibinfo{person}{Ruixiang Xue}, \bibinfo{person}{Jiaxin Li}, \bibinfo{person}{Dandan Ding}, \bibinfo{person}{Yi Lin}, {and} \bibinfo{person}{Zhan Ma}.} \bibinfo{year}{2024}\natexlab{b}.
\newblock \showarticletitle{A versatile point cloud compressor using universal multiscale conditional coding--Part I: Geometry}.
\newblock \bibinfo{journal}{\emph{IEEE transactions on pattern analysis and machine intelligence}} (\bibinfo{year}{2024}).
\newblock


\bibitem[Wang et~al\mbox{.}(2021)]%
        {wang2021lossy}
\bibfield{author}{\bibinfo{person}{Jianqiang Wang}, \bibinfo{person}{Hao Zhu}, \bibinfo{person}{Haojie Liu}, {and} \bibinfo{person}{Zhan Ma}.} \bibinfo{year}{2021}\natexlab{}.
\newblock \showarticletitle{Lossy point cloud geometry compression via end-to-end learning}.
\newblock \bibinfo{journal}{\emph{IEEE Transactions on Circuits and Systems for Video Technology}} \bibinfo{volume}{31}, \bibinfo{number}{12} (\bibinfo{year}{2021}), \bibinfo{pages}{4909--4923}.
\newblock


\bibitem[Zerman et~al\mbox{.}(2020)]%
        {zerman2020textured}
\bibfield{author}{\bibinfo{person}{Emin Zerman}, \bibinfo{person}{Cagri Ozcinar}, \bibinfo{person}{Pan Gao}, {and} \bibinfo{person}{Aljosa Smolic}.} \bibinfo{year}{2020}\natexlab{}.
\newblock \showarticletitle{Textured mesh vs coloured point cloud: A subjective study for volumetric video compression}. In \bibinfo{booktitle}{\emph{2020 Twelfth International Conference on Quality of Multimedia Experience (QoMEX)}}. IEEE, \bibinfo{pages}{1--6}.
\newblock


\bibitem[Zhang et~al\mbox{.}(2022)]%
        {zhang2022yuzu}
\bibfield{author}{\bibinfo{person}{Anlan Zhang}, \bibinfo{person}{Chendong Wang}, \bibinfo{person}{Bo Han}, {and} \bibinfo{person}{Feng Qian}.} \bibinfo{year}{2022}\natexlab{}.
\newblock \showarticletitle{YuZu: Neural-Enhanced volumetric video streaming}. In \bibinfo{booktitle}{\emph{19th USENIX Symposium on Networked Systems Design and Implementation (NSDI 22)}}. \bibinfo{pages}{137--154}.
\newblock


\bibitem[Zhang et~al\mbox{.}(2023)]%
        {zhang2023yoga}
\bibfield{author}{\bibinfo{person}{Junteng Zhang}, \bibinfo{person}{Tong Chen}, \bibinfo{person}{Dandan Ding}, {and} \bibinfo{person}{Zhan Ma}.} \bibinfo{year}{2023}\natexlab{}.
\newblock \showarticletitle{YOGA: Yet Another Geometry-based Point Cloud Compressor}. In \bibinfo{booktitle}{\emph{Proceedings of the 31st ACM International Conference on Multimedia}}.
\newblock


\bibitem[Zhang et~al\mbox{.}(2018)]%
        {zhang2018unreasonable}
\bibfield{author}{\bibinfo{person}{Richard Zhang}, \bibinfo{person}{Phillip Isola}, \bibinfo{person}{Alexei~A Efros}, \bibinfo{person}{Eli Shechtman}, {and} \bibinfo{person}{Oliver Wang}.} \bibinfo{year}{2018}\natexlab{}.
\newblock \showarticletitle{The unreasonable effectiveness of deep features as a perceptual metric}. In \bibinfo{booktitle}{\emph{CVPR}}. \bibinfo{pages}{586--595}.
\newblock


\bibitem[Zhou et~al\mbox{.}(2024)]%
        {zhou2024cap}
\bibfield{author}{\bibinfo{person}{Junsheng Zhou}, \bibinfo{person}{Baorui Ma}, \bibinfo{person}{Shujuan Li}, \bibinfo{person}{Yu-Shen Liu}, \bibinfo{person}{Yi Fang}, {and} \bibinfo{person}{Zhizhong Han}.} \bibinfo{year}{2024}\natexlab{}.
\newblock \showarticletitle{CAP-UDF: Learning unsigned distance functions progressively from raw point clouds with consistency-aware field optimization}.
\newblock \bibinfo{journal}{\emph{IEEE Transactions on Pattern Analysis and Machine Intelligence}} (\bibinfo{year}{2024}).
\newblock


\bibitem[Zhou et~al\mbox{.}(2018)]%
        {zhou2018open3d}
\bibfield{author}{\bibinfo{person}{Qian-Yi Zhou}, \bibinfo{person}{Jaesik Park}, {and} \bibinfo{person}{Vladlen Koltun}.} \bibinfo{year}{2018}\natexlab{}.
\newblock \showarticletitle{Open3D: A modern library for 3D data processing}.
\newblock \bibinfo{journal}{\emph{arXiv preprint arXiv:1801.09847}} (\bibinfo{year}{2018}).
\newblock


\bibitem[Zwicker et~al\mbox{.}(2001a)]%
        {zwicker2001ewa}
\bibfield{author}{\bibinfo{person}{Matthias Zwicker}, \bibinfo{person}{Hanspeter Pfister}, \bibinfo{person}{Jeroen Van~Baar}, {and} \bibinfo{person}{Markus Gross}.} \bibinfo{year}{2001}\natexlab{a}.
\newblock \showarticletitle{EWA volume splatting}. In \bibinfo{booktitle}{\emph{Proceedings Visualization, 2001. VIS'01.}} IEEE, \bibinfo{pages}{29--538}.
\newblock


\bibitem[Zwicker et~al\mbox{.}(2001b)]%
        {zwicker2001surface}
\bibfield{author}{\bibinfo{person}{Matthias Zwicker}, \bibinfo{person}{Hanspeter Pfister}, \bibinfo{person}{Jeroen Van~Baar}, {and} \bibinfo{person}{Markus Gross}.} \bibinfo{year}{2001}\natexlab{b}.
\newblock \showarticletitle{Surface splatting}. In \bibinfo{booktitle}{\emph{Proceedings of the 28th annual conference on Computer graphics and interactive techniques}}. \bibinfo{pages}{371--378}.
\newblock


\bibitem[Zwicker et~al\mbox{.}(2004)]%
        {zwicker2004perspective}
\bibfield{author}{\bibinfo{person}{Matthias Zwicker}, \bibinfo{person}{Jussi Rasanen}, \bibinfo{person}{Mario Botsch}, \bibinfo{person}{Carsten Dachsbacher}, {and} \bibinfo{person}{Mark Pauly}.} \bibinfo{year}{2004}\natexlab{}.
\newblock \showarticletitle{Perspective accurate splatting}. In \bibinfo{booktitle}{\emph{Proceedings-Graphics Interface}}. \bibinfo{pages}{247--254}.
\newblock


\end{thebibliography}

% %%
% %% If your work has an appendix, this is the place to put it.
% \appendix

% \section{Research Methods}

% \subsection{Part One}

% Lorem ipsum dolor sit amet, consectetur adipiscing elit. Morbi
% malesuada, quam in pulvinar varius, metus nunc fermentum urna, id
% sollicitudin purus odio sit amet enim. Aliquam ullamcorper eu ipsum
% vel mollis. Curabitur quis dictum nisl. Phasellus vel semper risus, et
% lacinia dolor. Integer ultricies commodo sem nec semper.

% \subsection{Part Two}

% Etiam commodo feugiat nisl pulvinar pellentesque. Etiam auctor sodales
% ligula, non varius nibh pulvinar semper. Suspendisse nec lectus non
% ipsum convallis congue hendrerit vitae sapien. Donec at laoreet
% eros. Vivamus non purus placerat, scelerisque diam eu, cursus
% ante. Etiam aliquam tortor auctor efficitur mattis.

% \section{Online Resources}

% Nam id fermentum dui. Suspendisse sagittis tortor a nulla mollis, in
% pulvinar ex pretium. Sed interdum orci quis metus euismod, et sagittis
% enim maximus. Vestibulum gravida massa ut felis suscipit
% congue. Quisque mattis elit a risus ultrices commodo venenatis eget
% dui. Etiam sagittis eleifend elementum.

% Nam interdum magna at lectus dignissim, ac dignissim lorem
% rhoncus. Maecenas eu arcu ac neque placerat aliquam. Nunc pulvinar
% massa et mattis lacinia.

\end{document}